%% file: acl2023.tex
\pdfoutput=1
\documentclass[11pt]{article}

\usepackage{ACL2023}

\usepackage{times}
\usepackage{latexsym}

\usepackage[T1]{fontenc}
\usepackage[utf8]{inputenc}

\usepackage{microtype}

\usepackage{inconsolata}
\usepackage{amsmath}
\usepackage{bm}
\usepackage{url}
\usepackage{booktabs}
\usepackage{graphicx}
\usepackage{multirow}
\usepackage{multicol}
\usepackage{amssymb}
\usepackage{makecell}
\usepackage[ruled,linesnumbered,commentsnumbered]{algorithm2e}

\title{Understanding and Bridging the Modality Gap for Speech Translation}

\author{
    Qingkai Fang$^{1,2}$,
    Yang Feng$^{1,2}$\thanks{ $\;\;$Corresponding author: Yang Feng.} \\
    \textsuperscript{\rm1}Key Laboratory of Intelligent Information Processing \\ Institute of Computing Technology, Chinese Academy of Sciences (ICT/CAS) \\
    \textsuperscript{\rm2}University of Chinese Academy of Sciences, Beijing, China \\
    \texttt{\{fangqingkai21b, fengyang\}@ict.ac.cn} \\
}

\begin{document}
\maketitle
\begin{abstract}
\input{Sections/000_abstract}
\end{abstract}

\section{Introduction}
\input{Sections/100_introduction}

\section{Background}
\input{Sections/200_background}

\section{Preliminary Studies on Modality Gap}
\input{Sections/300_preliminary}

\section{Method: \textsc{Cress}}
\input{Sections/400_method}

\section{Experiments}
\input{Sections/500_experiment}

\section{Analysis and Discussion}
\input{Sections/600_analysis_discussion}

\section{Related Work}
\input{Sections/700_related_work}

\section{Conclusion}
\input{Sections/800_conclusion}

\section*{Limitations}
\input{Sections/X00_limitations}

\section*{Ethics Statement}
\input{Sections/X00_ethics_statement}

\section*{Acknowledgements}
\input{Sections/X00_acknowledgement}

\bibliography{anthology,custom}
\bibliographystyle{acl_natbib}

\clearpage
\newpage
\appendix

\section{Statistics of all datasets}
\input{Sections/A00_statistics}

% \section{Impact of Base and Scale Parameters in Token-level Adaptive Training}
% \input{Sections/H00_base_and_scale.tex}

% \section{The Choice of Decay Parameter in Scheduled Sampling}
% \input{Sections/D00_impact_of_decay_parameter}

\section{Impact of Different Acoustic Encoders}
\input{Sections/C00_impact_of_acoustic_encoder}

\section{Results on CoVoST 2 En$\rightarrow$De}
\input{Sections/G00_covost2}

\section{Discussion about the Training Speed}
\input{Sections/B00_discussion_training_cost}

% \section{Performance of MT Task}
% \input{Sections/F00_mt_performance}

\section{ChrF++ Scores on MuST-C Dataset}
\input{Sections/E00_results_on_chrf}

\end{document}

%% file: Sections/000_abstract.tex
How to achieve better end-to-end speech translation (ST) by leveraging (text) machine translation (MT) data? Among various existing techniques, multi-task learning is one of the effective ways to share knowledge between ST and MT in which additional MT data can help to learn source-to-target mapping. However, due to the differences between speech and text, there is always a gap between ST and MT. In this paper, we first aim to understand this \emph{modality gap} from the target-side representation differences, and link the modality gap to another well-known problem in neural machine translation: exposure bias. We find that the modality gap is relatively small during training except for some difficult cases, but keeps increasing during inference due to the cascading effect. To address these problems, we propose the \textbf{C}ross-modal \textbf{Re}gularization with \textbf{S}cheduled \textbf{S}ampling (\textbf{\textsc{Cress}}) method. Specifically, we regularize the output predictions of ST and MT, whose target-side contexts are derived by sampling between ground truth words and self-generated words with a varying probability. Furthermore, we introduce token-level adaptive training which assigns different training weights to target tokens to handle difficult cases with large modality gaps. Experiments and analysis show that our approach effectively bridges the modality gap, and achieves promising results in all eight directions of the MuST-C dataset.\footnote{Code is publicly available at \url{https://github.com/ictnlp/CRESS}.}

%% file: Sections/100_introduction.tex
End-to-end speech translation (ST) aims to translate speech signals to text in another language directly. Compared to traditional cascaded methods, which combine automatic speech recognition (ASR) and machine translation (MT) models in a pipeline manner, end-to-end ST could avoid error propagation and high latency \citep{sperber-paulik-2020-speech}. Recently, end-to-end ST models have achieved comparable or even better results than cascaded ST models \citep{bentivogli-etal-2021-cascade, anastasopoulos-etal-2021-findings, anastasopoulos-etal-2022-findings}.

However, due to the scarcity of ST data, it is difficult to directly learn a mapping from source speech to the target text. Previous works often leverage MT data to help the training with multi-task learning \citep{ye-etal-2022-cross, tang-etal-2021-improving}. 
% By sharing encoder and decoder between ST and MT, multi-task learning enables knowledge sharing across tasks and tends to learn similar representations from different modalities.
By sharing encoder and decoder between ST and MT, the model tends to learn similar representations from different modalities. 
In this way, the auxiliary MT task can help build the source-to-target mapping. However, there remains a gap between ST and MT due to the differences between speech and text. In this paper, we measure the \emph{modality gap} with representation differences of the last decoder layer between ST and MT, because the representation of this layer will be mapped into the embedding space to obtain the final translation. A significant modality gap potentially causes different predictions, which makes ST lag behind MT.

% However, it is difficult to learn a mapping from source speech to target text directly with limited ST data. Previous work often leverage MT data to help the training, with techniques like multi-task learning \citep{ye-etal-2022-cross, tang-etal-2021-improving, fang-etal-2022-stemm}, which enables knowledge sharing between ST and MT. However, there remains a gap between ST and MT due to the differences between speech and text, as well as the complexity and diversity of speech itself. Therefore, in this paper, we aim to first \emph{understand} this gap, \emph{i.e.}, the modality gap between ST and MT, and then propose methods to \emph{bridge} the modality gap.

% The predictions of the translation model depend only on the representations of the last decoder layer. In this way, we measure the modality gap with the cosine similarity between the last decoder layer representations of ST and MT.

Thanks to multi-task learning, we observe that when training with teacher forcing, where both ST and MT use ground truth words as target-side contexts, the modality gap is relatively small except for some difficult cases. However, the exposure bias problem can make things worse. During inference, both ST and MT predict the next token conditioned on their previously generated tokens, which may be different due to the modality gap. Moreover, different predictions at the current step may lead to even more different predictions at the next step. As a result, the modality gap will increase step by step due to this cascading effect.

To solve these problems, we propose the \textbf{C}ross-modal \textbf{Re}gularization with \textbf{S}cheduled \textbf{S}ampling (\textbf{\textsc{Cress}}) method. To reduce the effect of exposure bias, we introduce scheduled sampling during training, where the target-side contexts are sampled between ground truth words and self-generated words with a changing probability. Based on this, we propose to regularize ST and MT in the output space to bridge the modality gap by minimizing the Kullback-Leibler (KL) divergence between their predictions. This will encourage greater consistency between ST and MT predictions based on partial self-generated words, which is closer to the inference mode. Besides, to handle the difficult cases, we introduce token-level adaptive training for \textbf{\textsc{Cress}}, where each target token is given a varying weight during training according to the scale of the modality gap. In this way, those cases with significant modality gaps will be emphasized. We conduct experiments on the ST benchmark dataset MuST-C \citep{di-gangi-etal-2019-must}. Results show that our approach significantly outperforms the strong multi-task learning baseline, with 1.8 BLEU improvements in the base setting and 1.3 BLEU improvements in the expanded setting on average. Further analysis shows that our approach effectively bridges the modality gap and improves the translation quality, especially for long sentences. 

% In the following of this paper, we first introduce the background of end-to-end speech translation in Section \ref{sec:background}. In Section \ref{sec:preliminary}, we conduct some preliminary studies to \emph{understand} the modality gap, and show its connection to exposure bias. In Section \ref{sec:method}, we introduce our proposed \textbf{\textsc{Cress}} method, which aims to \emph{bridge} the modality gap. Section \ref{sec:experiments} and \ref{sec:analysis} show the experiments and analysis of our approach.

%% file: Sections/200_background.tex
\label{sec:background}
\subsection{End-to-end Speech Translation}
\input{Sections/210_e2est}
\subsection{Multi-task Learning for ST}
\input{Sections/220_multi_task_st}

%% file: Sections/210_e2est.tex
End-to-end speech translation (ST) directly translates speech in the source language to text in the target language. The corpus of ST is usually composed of triplet data $\mathcal{D}=\{(\mathbf{s}, \mathbf{x}, \mathbf{y})\}$. Here $\mathbf{s}=(s_1, ..., s_{|\mathbf{s}|})$ is the sequence of audio wave, $\mathbf{x}=(x_1,...,x_{|\mathbf{x}|})$ is the transcription and $\mathbf{y}=(y_1,...,y_{|\mathbf{y}|})$ is the translation. Similar to previous work \citep{ye2021end, fang-etal-2022-stemm}, our ST model is composed of an acoustic encoder and a translation model. The acoustic encoder is a pre-trained HuBERT \citep{hubert} model followed by two convolutional layers, which are used to reduce the length of the speech sequence. The translation model follows standard Transformer \citep{transformer} encoder-decoder architecture, where the encoder contains $N$ Transformer encoder layers, and the decoder contains $N$ Transformer decoder layers. We first pre-train the translation model with MT data, and then optimize the whole model by minimizing a cross-entropy loss:
% The translation model is first pre-trained with text translation data. The whole model is optimized by minimizing a cross-entropy loss:
\begin{gather}
    \mathcal{L}_{\rm ST} = -\sum_{i=1}^{|\mathbf{y}|}\log p(y_i|\mathbf{s}, \mathbf{y}_{<i}), \\
    p(y_i|\mathbf{s}, \mathbf{y}_{<i})\propto\exp(\mathbf{W}\cdot f(\mathbf{s}, \mathbf{y}_{<i})), \label{eq:st_prob}
\end{gather}
where $f$ is a mapping from the input speech $\mathbf{s}$ and target prefix $\mathbf{y}_{<i}$ to the representation of the last decoder layer at step $i$. $\mathbf{W}$ is used to transform the dimension to the size of the target vocabulary.

%% file: Sections/220_multi_task_st.tex
\label{sec:mtl}
Multi-task learning (MTL) has been proven useful for sharing knowledge between text translation and speech translation \citep{tang-etal-2021-improving}, where an auxiliary MT task is introduced during training:
\begin{gather}
    \mathcal{L}_{\rm MT} = -\sum_{i=1}^{|\mathbf{y}|}\log p(y_i|\mathbf{x}, \mathbf{y}_{<i}), \\
    p(y_i|\mathbf{x}, \mathbf{y}_{<i})\propto\exp(\mathbf{W}\cdot f(\mathbf{x}, \mathbf{y}_{<i})).
\end{gather}
Note that both modalities (i.e., speech and text) share all transformer encoder and decoder layers. Finally, the training objective is written as follows:
\begin{equation}
    \mathcal{L}_{\rm MTL} = \mathcal{L}_{\rm ST} + \mathcal{L}_{\rm MT}.
\end{equation}

%% file: Sections/300_preliminary.tex
\label{sec:preliminary}
With multi-task learning, most of the knowledge of MT can be transferred to ST. However, the performance gap between ST and MT still exists. In this section, we first conduct some preliminary studies with our multi-task learning baseline model to understand where this gap comes from. 

\subsection{Definition of the Modality Gap}
\input{Sections/310_token_level_modality_gap}

\input{Figures/figure_cos_dist}

\subsection{Connection between Exposure Bias and Modality Gap}
\input{Sections/320_exposure_bias}

%% file: Sections/310_token_level_modality_gap.tex
The gap between ST and MT is related to the prediction difference at each decoding step, while the prediction depends only on the representation of the last decoder layer. Therefore, we define the \emph{modality gap} at the $i$-th decoding step as follows:
\begin{equation}
    G(\mathbf{s}, \mathbf{y}_{<i} \| \mathbf{x}, \mathbf{y}_{<i})=1-cos(f(\mathbf{s}, \mathbf{y}_{<i}), f(\mathbf{x}, \mathbf{y}_{<i})),
\end{equation}
where $cos$ is the cosine similarity function $cos(\mathbf{a}, \mathbf{b})=\mathbf{a}^\top\mathbf{b}/\|\mathbf{a}\|\|\mathbf{b}\|$. A larger cosine similarity indicates a smaller modality gap.

To understand the extent of the modality gap, we count the distribution of $G(\mathbf{s}, \mathbf{y}_{<i} \| \mathbf{x}, \mathbf{y}_{<i})$ based on all triples $(\mathbf{s}, \mathbf{x}, \mathbf{y}_{<i})$ in the MuST-C \citep{di-gangi-etal-2019-must} En$\rightarrow$De \texttt{dev} set. As shown in Figure \ref{fig:cos_dist}, the modality gap is relatively small ($<10\%$) in most cases, which proves the effectiveness of multi-task learning in sharing knowledge across ST and MT. However, we also observe a long-tail problem: there is a large difference between ST and MT representations in some difficult cases.

%% file: Figures/figure_cos_dist.tex
\begin{figure}[t]
    \centering
    \includegraphics[width=\linewidth]{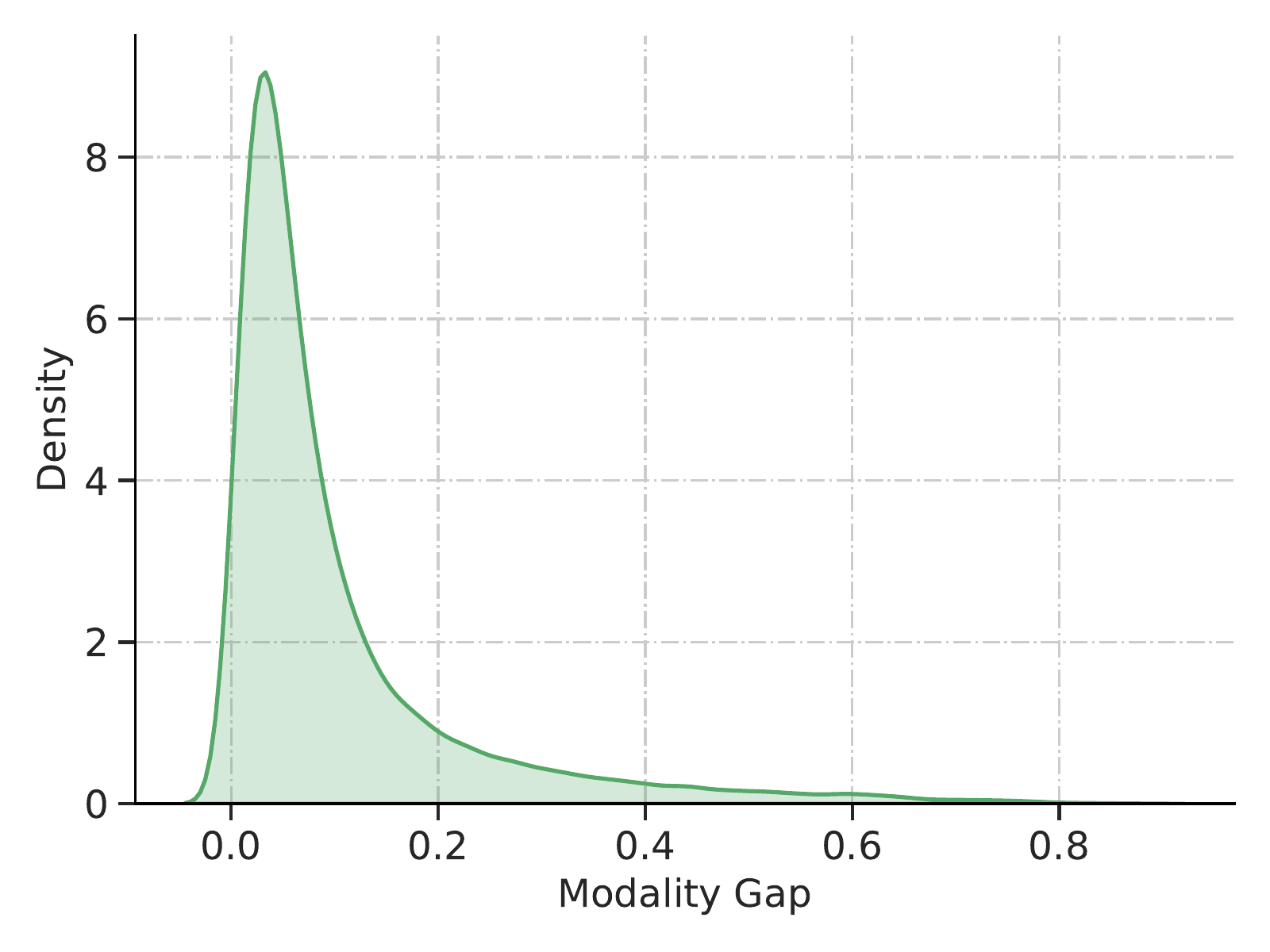}
    \caption{Distribution of the modality gap on MuST-C En$\rightarrow$De \texttt{dev} set with kernel density estimation (KDE).}
    \label{fig:cos_dist}
\end{figure}

%% file: Sections/320_exposure_bias.tex
\emph{Exposure bias}, a discrepancy between training and inference, is a well-known problem in neural machine translation \citep{schedule-sampling, sequence-level-training, wang-sennrich-2020-exposure, arora-etal-2022-exposure}. During training with \emph{teacher forcing}, both ST and MT predict the next token conditioned on the ground truth target prefix $\mathbf{y}_{<i}$. However, during inference, the predictions of ST and MT depend on their previously generated tokens by the model itself (denoted as $\widehat{\mathbf{y}}^{s}_{<i}$ and $\widehat{\mathbf{y}}^{x}_{<i}$ for ST and MT respectively), which might be different due to the modality gap. Furthermore, different predictions at the current decoding step result in different target prefixes for ST and MT, potentially causing even more different predictions at the next step. Such cascading effect will enlarge the modality gap step by step during inference.

To prove our hypothesis, we present the curves of the modality gap with decoding steps under \emph{teacher forcing}, \emph{beam search}, and \emph{greedy search} strategies, respectively. As shown in Figure \ref{fig:exposure_bias}, with \emph{teacher forcing}, there is no significant difference in the modality gap across steps, as both ST and MT depend on the same target prefix at any step. Hence, the modality gap $G(\mathbf{s}, \mathbf{y}_{<i} \| \mathbf{x}, \mathbf{y}_{<i})$ only comes from the difference between input speech $\mathbf{s}$ and text $\mathbf{x}$. However, when decoding with \emph{greedy search}, due to the cascading effect mentioned above, the self-generated target prefix $\widehat{\mathbf{y}}^{s}_{<i}$ and $\widehat{\mathbf{y}}^{x}_{<i}$ become increasingly different, making the modality gap $G(\mathbf{s}, \widehat{\mathbf{y}}^{s}_{<i} \| \mathbf{x}, \widehat{\mathbf{y}}^{x}_{<i})$ keep increasing with decoding steps. A simple way to alleviate this problem is \emph{beam search}, which considers several candidate tokens rather than a single one at each decoding step. When there is an overlap between candidate tokens of ST and MT, the cascading effect will be reduced, thus slowing down the increase of the modality gap. 

\input{Figures/figure_exposure_bias}

%% file: Figures/figure_exposure_bias.tex
\begin{figure}[t]
    \centering
    \includegraphics[width=\linewidth]{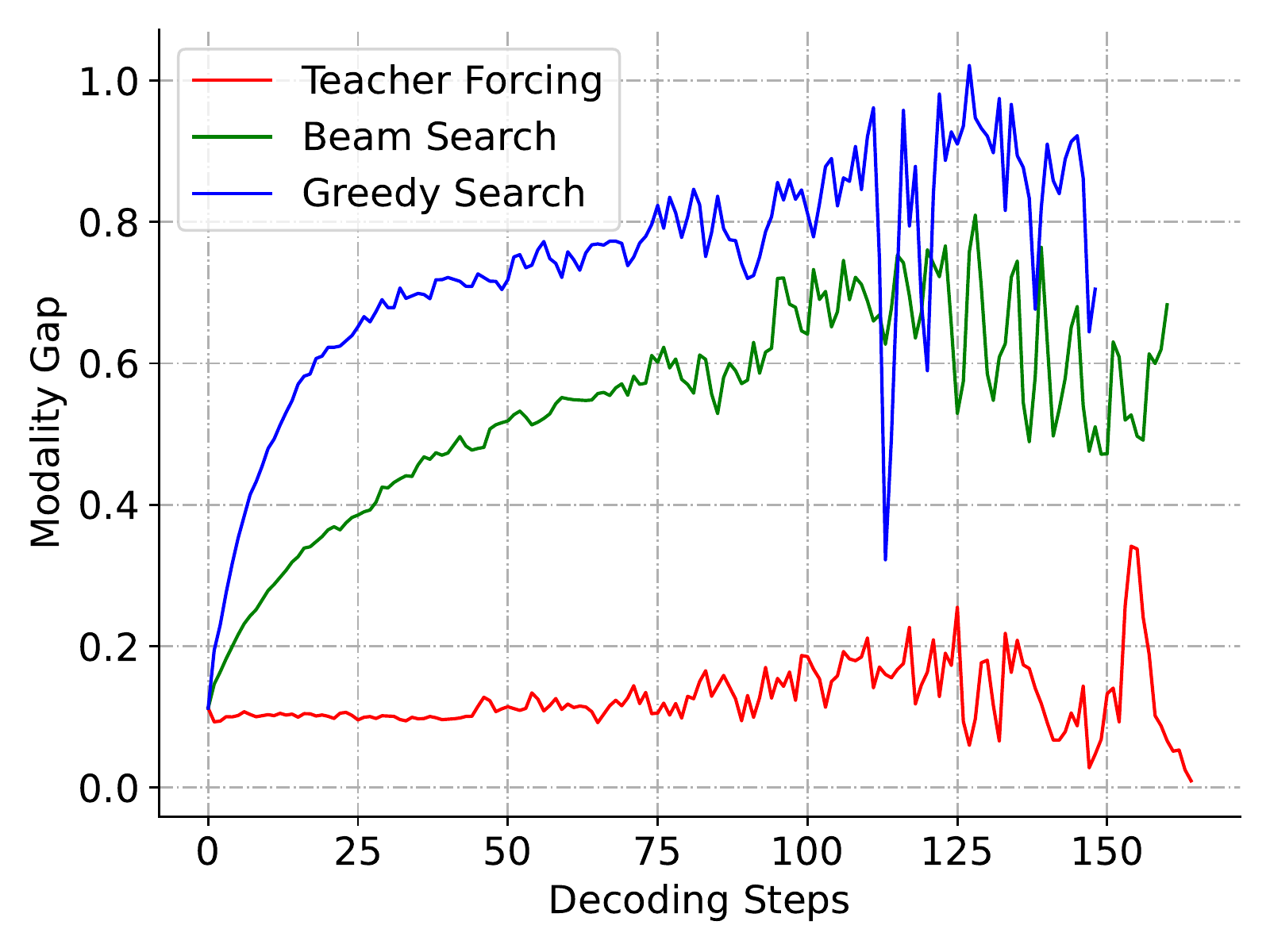}
    \caption{Curves of the average modality gap on MuST-C En$\rightarrow$De \texttt{dev} set with decoding steps under \emph{teacher forcing}, \emph{beam search}, and \emph{greedy search} strategies. For \emph{beam search}, we have several candidate translations. The modality gap is calculated with the average representation of all candidates. We set a beam size of 8.}
    \label{fig:exposure_bias}
\end{figure}

%% file: Sections/400_method.tex
\label{sec:method}
Our preliminary studies in Section \ref{sec:preliminary} show that:
\begin{itemize}
    \item The modality gap will be enlarged during inference due to exposure bias.
    \item The modality gap may be significant in some difficult cases.
\end{itemize}

Inspired by these, we propose the \textbf{C}ross-modal \textbf{Re}gularization with \textbf{S}cheduled \textbf{S}ampling (\textbf{\textsc{Cress}}) method to bridge the modality gap, especially in inference mode (Section \ref{sec:cress}). Furthermore, we propose a token-level adaptive training method for \textbf{\textsc{Cress}} to handle difficult cases (Section \ref{sec:adaptive}). 
\subsection{Cross-modal Regularization with Scheduled Sampling (\textsc{Cress})}
\input{Figures/figure_method}
\input{Sections/410_schedule_sampling}
\subsection{Token-level Adaptive Training for \textsc{Cress}}
\input{Sections/420_adaptive_training}

%% file: Figures/figure_method.tex
\begin{figure*}[t]
    \centering
    \includegraphics[width=\textwidth]{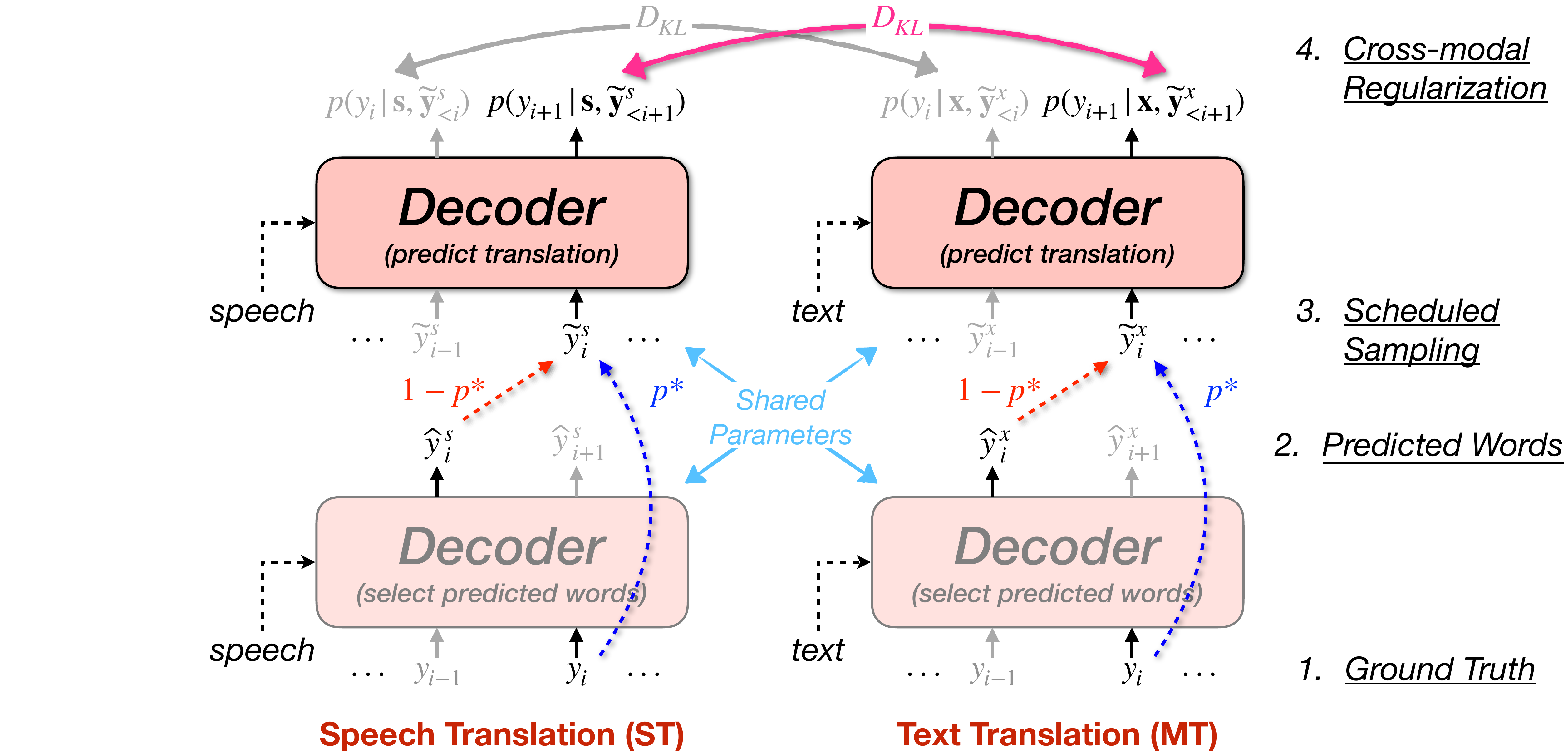}
    \caption{Overview of our proposed \textbf{\textsc{Cress}}. Note that the step of selecting predicted words has no gradient calculation and is fully parallelized.}
    \label{fig:method}
\end{figure*}

%% file: Sections/410_schedule_sampling.tex
\label{sec:cress}
To bridge the modality gap during inference, we adopt scheduled sampling for both ST and MT to approximate the inference mode at training time. After that, we add a regularization loss between the predictions of ST and MT based on the part of their self-generated words as context. This allows for more consistent predictions between ST and MT during inference, thus reducing the performance gap between ST and MT. Figure \ref{fig:method} illustrates the main framework of our method.

\paragraph{Scheduled Sampling}
\emph{Scheduled sampling} \citep{schedule-sampling}, which samples between ground truth words and self-generated words, i.e., \emph{predicted words}, with a certain probability as target-side context, has proven helpful in alleviating exposure bias. In general, the input at the $\{i+1\}$-th decoding step should be the ground truth word $y_{i}$ during training. With scheduled sampling, it can also be substituted by a predicted word. Next, we describe how to select the predicted word $\widehat{y}_{i}^{s}$ for ST and $\widehat{y}_{i}^{x}$ for MT. For ST, we follow \citet{zhang-etal-2019-bridging} to select the predicted word $\widehat{y}_{i}^{s}$ by sampling from the word distribution $p(y_i|\mathbf{s}, \mathbf{y}_{<i})$ in Equation (\ref{eq:st_prob}) with \emph{Gumbel-Max} technique \citep{gumbel1954statistical, asampling}, a method to draw a sample from a categorical distribution:
\begin{gather}
    \eta=-\log(-\log u), \\
    \widehat{y}_{i}^{s}=\arg\max\left( \mathbf{W}\cdot f(\mathbf{s}, \mathbf{y}_{<i})+\eta \right),
\end{gather}
where $\eta$ is the Gumbel noise calculated from the uniform noise $u\sim \mathcal{U}(0,1)$. Similarly, for MT, there is:
\begin{equation}
    \widehat{y}_{i}^{x}=\arg\max\left( \mathbf{W}\cdot f(\mathbf{x}, \mathbf{y}_{<i})+\eta \right).
\end{equation}
Note that we may omit the superscript and denote the predicted word for both ST and MT by $\widehat{y}_i$ in the following.

How to select between the ground truth word $y_{i}$ and the predicted word $\widehat{y}_{i}$? Similar to \citet{schedule-sampling, zhang-etal-2019-bridging}, we randomly sample from both with a varying probability. We denote the probability of selecting from the ground truth word as $p^*$. At the beginning of training, since the model is not yet well trained, we select more from the ground truth words (with larger $p^*$) to help the model converge. In the later stages of training, we select more from the predicted words (with smaller $p^*$), which is closer to the situation during inference. To achieve this, we decrease $p^*$ with a function of the index of training epochs $e$:
\begin{equation}
    p^*=\frac{\mu}{\mu+\exp(e/\mu)},\label{eq:decay}
\end{equation}
where $\mu$ is a hyper-parameter. With scheduled sampling, the target-side context becomes $\widetilde{\mathbf{y}}=(\widetilde{y}_1,...,\widetilde{y}_{|\mathbf{y}|})$, where
\begin{equation}
    \widetilde{y}_i=\begin{cases}
        y_i, & p \leq p^* \\
        \widehat{y}_i, & p > p^*
    \end{cases},
\end{equation}
where $p$ is sampled from the uniform distribution $\mathcal{U}(0, 1)$. Using $\widetilde{\mathbf{y}}^s$ and $\widetilde{\mathbf{y}}^x$ to denote the target-side context of ST and MT respectively, the loss functions of ST and MT become:
\begin{align}
    \mathcal{L}_{\rm ST}^{\textsc{Cress}} &= -\sum_{i=1}^{|\mathbf{y}|}\log p(y_i|\mathbf{s}, \widetilde{\mathbf{y}}^s_{<i}), \label{eq:st_cress} \\
    \mathcal{L}_{\rm MT}^{\textsc{Cress}} &= -\sum_{i=1}^{|\mathbf{y}|}\log p(y_i|\mathbf{x}, \widetilde{\mathbf{y}}^x_{<i}), \label{eq:mt_cress}
\end{align}

\paragraph{Cross-modal Regularization}
To bridge the modality gap in inference mode, we expect the predictions of ST and MT with scheduled sampling to be consistent. Inspired by recent works of consistency training \citep{rdrop, guo-etal-2022-prediction-difference}, we regularize ST and MT in the output space. Specifically, we minimize the bidirectional Kullback-Leibler (KL) divergence between the output distributions of ST and MT at each step:
\begin{equation}
    \begin{split}
    \mathcal{L}_{\rm Reg}^{\textsc{Cress}} = \sum_{i=1}^{|\mathbf{y}|} & \frac{1}{2}(\mathcal{D}_{\rm KL}(p(y_i|\mathbf{s}, \widetilde{\mathbf{y}}^s_{<i}) \| p(y_i|\mathbf{x}, \widetilde{\mathbf{y}}^x_{<i})) \\
    & + \mathcal{D}_{\rm KL}(p(y_i|\mathbf{x}, \widetilde{\mathbf{y}}^x_{<i}) \| p(y_i|\mathbf{s}, \widetilde{\mathbf{y}}^s_{<i}))). \label{eq:reg_cress}
    \end{split}
\end{equation}

With the translation loss in Equation (\ref{eq:st_cress}) and (\ref{eq:mt_cress}), the final training objective is:
\begin{equation}
    \mathcal{L}^{\textsc{Cress}} = \mathcal{L}_{\rm ST}^{\textsc{Cress}} + \mathcal{L}_{\rm MT}^{\textsc{Cress}} + \lambda \mathcal{L}_{\rm Reg}^{\textsc{Cress}},
\end{equation}
where $\lambda$ is the hyper-parameter to control the weight of $\mathcal{L}_{\rm Reg}^{\textsc{Cress}}$.

%% file: Sections/420_adaptive_training.tex
\label{sec:adaptive}
As mentioned above, the modality gap might be significant in some difficult cases. Inspired by the idea of token-level adaptive training \citep{gu-etal-2020-token, xu-etal-2021-bilingual, zhang-etal-2022-conditional}, we propose to treat each token adaptively according to the scale of the modality gap. The training objectives in Equation (\ref{eq:st_cress}), (\ref{eq:mt_cress}), and (\ref{eq:reg_cress}) are modified as follows:
\begin{gather}
    \mathcal{L}_{\rm ST}^{\textsc{Cress}} = -\sum_{i=1}^{|\mathbf{y}|}w_i\cdot\log p(y_i|\mathbf{s}, \widetilde{\mathbf{y}}^s_{<i}), \\
    \mathcal{L}_{\rm MT}^{\textsc{Cress}} = -\sum_{i=1}^{|\mathbf{y}|}w_i\cdot\log p(y_i|\mathbf{x}, \widetilde{\mathbf{y}}^x_{<i}), \\
    \begin{split}
    \mathcal{L}_{\rm Reg}^{\textsc{Cress}} = \sum_{i=1}^{|\mathbf{y}|} &\frac{1}{2}w_i(\mathcal{D}_{\rm KL}(p(y_i|\mathbf{s}, \widetilde{\mathbf{y}}^s_{<i}) \| p(y_i|\mathbf{x}, \widetilde{\mathbf{y}}^x_{<i})) \\
    &+ \mathcal{D}_{\rm KL}(p(y_i|\mathbf{x}, \widetilde{\mathbf{y}}^x_{<i}) \| p(y_i|\mathbf{s}, \widetilde{\mathbf{y}}^s_{<i}))),
    \end{split}
\end{gather}
where $w_i$ is the token-level weight defined by a linear function of the modality gap:
\begin{equation}
    w_i=B+S\cdot G(\mathbf{s}, \widetilde{\mathbf{y}}^s_{<i} \| \mathbf{x}, \widetilde{\mathbf{y}}^x_{<i}),
\end{equation}
where $B$ (base) and $S$ (scale) are hyper-parameters to control the lower bound and magnitude of change of $w_i$. In this way, cases with a large modality gap will be assigned a larger weight and thus emphasized during training. Note that the modality gap is computed on-the-fly during training.
% The final training objective with token-level adaptive training is as follows:
% \begin{equation}
%     \mathcal{L}^{\textsc{Cress+}} = \mathcal{L}_{\rm ST}^{\textsc{Cress+}} + \mathcal{L}_{\rm MT}^{\textsc{Cress+}} + \lambda \mathcal{L}_{\rm Reg}^{\textsc{Cress+}}.
% \end{equation}

%% file: Sections/500_experiment.tex
\label{sec:experiments}
\subsection{Datasets}
\input{Sections/510_datasets}
\subsection{Experimental Setups}
\input{Sections/520_setups}
\subsection{Main Results on MuST-C Dataset}
\input{Sections/530_results_mustc}

%% file: Sections/510_datasets.tex
\paragraph{ST Datasets}
We conduct experiments on MuST-C \citep{di-gangi-etal-2019-must} dataset, a multilingual ST dataset containing 8 translation directions: English (En) to German (De), French (Fr), Spanish (Es), Romanian (Ro), Russian (Ru), Italian (It), Portuguese (Pt) and Dutch (Nl). It contains at least 385 hours of TED talks with transcriptions and translations for each direction. We use \texttt{dev} set for validation and \texttt{tst-COMMON} set for evaluation.
\paragraph{External MT Datasets}
We also introduce external MT datasets to pre-train our translation model in the expanded setting. For En$\rightarrow$De/Fr/Es/Ro/Ru directions, we introduce data from WMT \citep{buck-koehn-2016-findings}. For En$\rightarrow$It/Pt/Nl, we introduce data from OPUS100\footnote{\url{http://opus.nlpl.eu/opus-100.php}} \citep{zhang-etal-2020-improving}. Table \ref{tab:data} in Appendix \ref{sec:statistics} lists the statistics of all datasets.

%% file: Sections/520_setups.tex
\input{Tables/table_main_results}
\label{sec:setup}
\paragraph{Pre-processing}
For \emph{speech} input, we use the raw 16-bit 16kHz mono-channel audio wave. For \emph{text} input, all sentences in ST and external MT datasets are tokenized and segmented into subwords using
SentencePiece\footnote{\url{https://github.com/google/sentencepiece}}. For each translation direction, the vocabulary is learned from the source and target texts from the ST dataset, with a size of 10K. For the external MT datasets, we filter out parallel sentence pairs whose length ratio exceeds 1.5.
\paragraph{Model Setting}
We use the pre-trained HuBERT model\footnote{\url{https://dl.fbaipublicfiles.com/hubert/hubert_base_ls960.pt}} to encode the input audio. Two 1-dimensional convolutional layers after HuBERT are set to kernel size 5, stride size 2, and padding 2. For the translation model, we employ Transformer architecture with the base configuration, which contains 6 encoder layers and 6 decoder layers, with 512 hidden states, 8 attention heads, and 2048 feed-forward hidden states for each layer. The translation model is first pre-trained with MT task using \emph{transcription-translation} pairs from the ST dataset (\textbf{base setting}), and also sentence pairs from the external MT dataset (\textbf{expanded setting}).

During MT pre-training, each batch has up to 33k text tokens. The maximum learning rate is set to 7e-4. During fine-tuning, each batch contains up to 16M audio frames. The maximum learning rate is set to 1e-4. We use Adam optimizer \citep{adam} with 4k warm-up steps. We set dropout to 0.1 and label smoothing to 0.1. The training will early stop if the BLEU score on the \texttt{dev} set did not increase for 10 epochs. During inference, we average the checkpoints of the last 10 epochs for evaluation. We use beam search with a beam size of 8. The length penalty is set to 1.2, 1.8, 0.6, 1.4, 0.8, 1.0, 1.4, and 1.0 for En$\rightarrow$De, Fr, Es, Ro, Ru, It, Pt and Nl, respectively. We use scareBLEU\footnote{\url{https://github.com/mjpost/sacrebleu}} \citep{post-2018-call} to compute case-sensitive detokenized BLEU \citep{papineni-etal-2002-bleu} scores and the statistical significance of translation results with paired bootstrap resampling\footnote{sacreBLEU signature: nrefs:1 | bs:1000 | seed:12345 | case:mixed | eff:no | tok:13a | smooth:exp | version:2.0.0}
\citep{koehn-2004-statistical}. 
% We also report ChrF++ scores in Appendix \ref{sec:chrf}. 
We implement our model with \emph{fairseq}\footnote{\url{https://github.com/pytorch/fairseq}} \citep{ott2019fairseq}. All models are trained on 4 Nvidia RTX 3090 GPUs.

For scheduled sampling, the decay parameter is set to $\mu=15$. For cross-modal regularization, the weight parameter is set to $\lambda=1.0$. For token-level adaptive training, we did a grid search for base and scale parameters on MuST-C En$\rightarrow$De \texttt{dev} set with $B\in\{0.6, 0.7, 0.8, 0.9, 1.0\}$ and $S\in\{0.05, 0.10, 0.20, 0.50, 1.00\}$. Finally, we set $B=0.7$ and $S=0.05$ for all translation directions. We start applying token-level adaptive training after the 20th epoch during training.

\paragraph{Baseline Systems}
We include several strong end-to-end ST models for comparison: Chimera \citep{han-etal-2021-learning}, XSTNet \citep{ye2021end}, STEMM \citep{fang-etal-2022-stemm}, ConST \citep{ye-etal-2022-cross}, STPT \citep{tang-etal-2022-unified}, and SpeechUT \citep{zhang2022speechut}. Besides, the multi-task learning baseline in Section \ref{sec:mtl} is also included as a strong baseline, which is denoted as \textbf{\textsc{MTL}}. We use \textbf{\textsc{Cress}} to denote our method with token-level adaptive training.

Among these models, Chimera, XSTNet, STEMM, and ConST combine pre-trained Wav2vec 2.0 \citep{baevski2020wav2vec} and pre-trained translation model together, and then fine-tune the whole model on ST datasets. Our implemented \textbf{\textsc{MTL}} and \textbf{\textsc{Cress}} follow a similar design, but we use HuBERT instead of Wav2vec 2.0 as we find HuBERT gives a stronger baseline (See Table \ref{tab:wav2vec} for details). STPT and SpeechUT jointly pre-train the model on speech and text data from scratch, which achieve better performance but also bring higher training costs\footnote{For example, the pre-training of SpeechUT takes 3 days with 32 V100 GPUs.}.

%% file: Tables/table_main_results.tex
\begin{table*}[t]
\centering
\resizebox{\linewidth}{!}{
\begin{tabular}{l|cccccccc|c}
\toprule
\multirow{2}{*}{\textbf{Models}} & \multicolumn{9}{c}{\textbf{BLEU}} \\
                        & En$\rightarrow$De & En$\rightarrow$Fr & En$\rightarrow$Es & En$\rightarrow$Ro & En$\rightarrow$Ru & En$\rightarrow$It & En$\rightarrow$Pt & En$\rightarrow$Nl & Avg. \\ 
\midrule
\multicolumn{10}{c}{\textbf{Base setting} (w/o external MT data)} \\
\midrule
% Fairseq ST \citep{wang2020fairseqs2t}  & 22.7 & 32.9 & 27.2 & 21.9 & 15.3 & 22.7 & 28.1 & 27.3 & 24.8 \\
% RevisitST \citep{revisit-st}           & 23.0 & 33.5 & 28.0 & 23.0 & 15.6 & 23.5 & 28.2 & 27.1 & 25.2 \\
% DDT \citep{le-etal-2020-dual}          & 23.6 & 33.5 & 28.1 & 22.9 & 15.2 & 24.2 & 30.0 & 27.6 & 25.6 \\
% TDA \citep{RegSTaaai}               & 25.4 & 36.1 & 29.6 & 23.9 & 16.4 & 25.1 & 31.1 & 29.6 & 27.2 \\
XSTNet \citep{ye2021end}               & 25.5 & 36.0 & 29.6 & 25.1 & 16.9 & 25.5 & 31.3 & 30.0 & 27.5 \\
STEMM \citep{fang-etal-2022-stemm}     & 25.6 & 36.1 & 30.3 & 24.3 & 17.1 & 25.6 & 31.0 & 30.1 & 27.5 \\
ConST \citep{ye-etal-2022-cross}       & 25.7 & 36.8 & 30.4 & 24.8 & 17.3 & 26.3 & 32.0 & 30.6 & 28.0 \\
\textbf{\textsc{MTL}}                  & 25.3 & 35.7 & 30.5 & 23.8    & 17.2    & 26.0    & 31.3    & 29.5    & 27.4    \\
\textbf{\textsc{Cress}}               & \textbf{27.2**} & \textbf{37.8**} & \textbf{31.9**} & \textbf{25.9**} & \textbf{18.7**}    & \textbf{27.3**}    & \textbf{33.0**}    & \textbf{31.6**}    & \textbf{29.2}    \\
\midrule
\multicolumn{10}{c}{\textbf{Expanded setting} (w/ external MT data)} \\
\midrule
% LAT \citep{le-etal-2021-lightweight}   & 24.7 & 35.0 & 28.7 & 23.8 & 16.4 & 25.0 & 31.1 & 28.8 & 26.7 \\
% TaskAware \citep{taskaware}            & 28.9 & -    & -    & -    & -    & -    & -    & -    & -    \\
Chimera \citep{han-etal-2021-learning} & 27.1 & 35.6 & 30.6 & 24.0 & 17.4 & 25.0 & 30.2 & 29.2 & 27.4 \\
XSTNet \citep{ye2021end}               & 27.1 & 38.0 & 30.8 & 25.7 & 18.5 & 26.4 & 32.4 & 31.2 & 28.8 \\
STEMM \citep{fang-etal-2022-stemm}     & 28.7 & 37.4 & 31.0 & 24.5 & 17.8 & 25.8 & 31.7 & 30.5 & 28.4 \\
ConST \citep{ye-etal-2022-cross}       & 28.3 & 38.3 & 32.0 & 25.6 & 18.9 & 27.2 & 33.1 & 31.7 & 29.4 \\
$^\dagger$STPT \citep{tang-etal-2022-unified}    & -    & 39.7 & 33.1 & -    & -    & -    & -    & -    & -    \\
$^\dagger$SpeechUT \citep{zhang2022speechut}    & \textbf{30.1} & \textbf{41.4} & \textbf{33.6} & -    & -    & -    & -    & -    & -    \\
\textbf{\textsc{MTL}}                  & 27.7 & 38.5 & 32.8 & 24.9    & 19.0    & 26.5    & 32.0    & 30.8    & 29.0    \\
\textbf{\textsc{Cress}}               & 29.4** & 40.1** & 33.2* & \textbf{26.4**}   & \textbf{19.7**}    & \textbf{27.6**}    & \textbf{33.6**}    & \textbf{32.3**}    &\textbf{30.3}    \\
\bottomrule
\end{tabular}}
\caption{BLEU scores on MuST-C \texttt{tst-COMMON} set. The external MT datasets are only used in the expanded setting. * and ** mean the improvements over \textbf{\textsc{MTL}} baseline are statistically significant ($p < 0.05$ and $p < 0.01$, respectively). $\dagger$: speech-text jointly pre-trained models whose training costs are much higher than our models.}
\label{tab:main}
\end{table*}

%% file: Sections/530_results_mustc.tex
\label{sec:results}
Table \ref{tab:main} shows the results on MuST-C \texttt{tst-COMMON} set in all eight directions. First, we find that our implemented \textbf{\textsc{MTL}} is a strong baseline compared with existing approaches. Second, our proposed \textbf{\textsc{Cress}} significantly outperforms \textbf{\textsc{MTL}} in both settings, with 1.8 BLEU improvement in the base setting and 1.3 BLEU improvement in the expanded setting on average, demonstrating the superiority of our approach. Besides, we report ChrF++ scores on MuST-C in Appendix \ref{sec:chrf}, and we also provide results on CoVoST 2 \citep{covost2} En$\rightarrow$De dataset in Appendix \ref{sec:covost}.

\input{Tables/table_ablation}

%% file: Tables/table_ablation.tex
\begin{table}[t]
\centering
\resizebox{\linewidth}{!}{
\begin{tabular}{c|ccc|c}
\toprule
    \textbf{\#} & \textbf{\makecell{Adaptive\\Training}} & \textbf{Regularization} & \textbf{\makecell{Scheduled\\Sampling}} & \textbf{BLEU} \\
    \midrule
    1 & \checkmark & \checkmark & \checkmark &  \textbf{29.4} \\
    2 & \texttimes & \checkmark & \checkmark &  29.0 \\
    3 & \texttimes & \checkmark & \texttimes &  28.4 \\
    4 & \texttimes & \texttimes & \checkmark &  28.0 \\
    5 & \checkmark & \texttimes & \texttimes &  27.5 \\
    6 & \texttimes & \texttimes & \texttimes &  27.7 \\
\bottomrule
\end{tabular}}
\caption{BLEU scores on MuST-C En$\rightarrow$De \texttt{tst-COMMON} set with different combinations of training techniques.}
% * and ** mean the improvements over \textbf{\textsc{MTL}} (\#5) are statistically significant ($p < 0.05$ and $p < 0.01$, respectively).}
\label{tab:ablation}
\end{table}

%% file: Sections/600_analysis_discussion.tex
\label{sec:analysis}

Results in Section \ref{sec:results} show the superiority of our method. To better understand \textbf{\textsc{Cress}}, we explore several questions in this section. All analysis experiments are conducted on MuST-C En$\rightarrow$De dataset in the expanded setting. 
% We also investigate the impact of different acoustic encoders (Wav2vec2.0 \citep{baevski2020wav2vec} and HuBERT) in Appendix \ref{sec:acoustic_encoder} and the impact of decay parameter in scheduled sampling in Appendix \ref{sec:decay}.

\textbf{(1) Do scheduled sampling, cross-modal regularization, and token-level adaptive training all matter?}
Scheduled sampling, regularization, and token-level adaptive training are effective techniques to improve the performance of translation models. To understand the role of each, we conduct ablation experiments in Table \ref{tab:ablation}. When only applying token-level adaptive training (\#5), we observe a performance decline of 0.2 BLEU since only adaptive training can not bridge the modality gap. When training with scheduled sampling only (\#4), we observe a slight improvement of 0.3 BLEU, probably due to the alleviation of exposure bias. When training with cross-modal regularization only (\#3), which encourages the consistency between predictions of ST and MT with ground truth target contexts, we observe an improvement of 0.7 BLEU. If we combine both (\#2), we obtain a much more significant boost of 1.3 BLEU, proving that both scheduled sampling and cross-modal regularization play a crucial role in our method. Furthermore, with token-level adaptive training (\#1), the improvement comes to 1.7 BLEU, which shows the benefit of treating different tokens differently according to the modality gap.

% \input{Figures/figure_two_minipage}
% \input{Figures/figure_heatmap}
\input{Figures/figure_cos_dist_cmp}
\input{Figures/figure_modality_gap_cmp}

\textbf{(2) Does \textsc{Cress} successfully bridge the modality gap?} To validate whether our approach successfully bridges the modality gap between ST and MT, we revisit the experiments in Section \ref{sec:preliminary}. Figure \ref{fig:cos_dist_cmp} shows the distribution of the modality gap with teacher forcing. We observe a general decrease in the modality gap compared with \textbf{\textsc{MTL}}. We also plot the curves of the modality gap with decoding steps of \textbf{\textsc{Cress}} under teacher forcing, greedy search, and beam search strategies. 
As shown in Figure \ref{fig:modality_gap_cmp}, our approach significantly slows down the increase of the modality gap compared with \textbf{\textsc{MTL}} baseline, suggesting that the predictions of ST and MT are more consistent during inference, demonstrating the effectiveness of our method in bridging the modality gap.

\textbf{(3) How base and scale hyper-parameters influence token-level adaptive training?}
$B$ (base) and $S$ (scale) are two important hyper-parameters in token-level adaptive training. We investigate how different combinations of $B$ and $S$ influence performance. As shown in Figure \ref{fig:heatmap}, token-level adaptive training can bring improvements in most cases. In particular, it usually performs better with smaller $B$ and smaller $S$, leading to a boost of up to 0.4 BLEU. We conclude that treating different tokens too differently is also undesirable. We use $B=0.7$ and $S=0.05$ for all translation directions.
\input{Figures/figure_heatmap}

% \input{Tables/table_wav2vec}

\textbf{(4) Does \textsc{Cress} successfully reduce the performance gap between ST and MT?}
As shown in Table \ref{tab:mt}, our method not only brings improvements to ST, but also gives a slight average boost of 0.3 BLEU to MT. We suggest that this may be due to the effect of regularization. More importantly, we find that the performance gap between ST and MT for \textbf{\textsc{Cress}} is significantly reduced compared to the \textbf{\textsc{MTL}} baseline (6.0$\rightarrow$5.0), which further demonstrates that the improvement in ST is mainly due to the effective reduction of the modality gap.

\input{Tables/table_mt}

\textbf{(5) Is \textsc{Cress} more effective for longer sentences?} The autoregressive model generates the translation step by step, making the translation of long sentences more challenging. We divide the MuST-C En$\rightarrow$De \texttt{dev} set into several groups according to the length of target sentences, and compute the BLEU scores in each group separately, as shown in Figure \ref{fig:bleu_sentence_length}. We observe that \textbf{\textsc{Cress}} achieve significant improvements over the baseline in all groups, especially for sentences longer than 45, which shows the superiority of our method when translating long sentences.

\textbf{(6) How the decay parameter in scheduled sampling influence the performance?}
In scheduled sampling, the probability of selecting the ground truth word $p^*$ keeps decreasing during training as the function in Equation (\ref{eq:decay}). Here, the hyper-parameter $\mu$ is used to control the shape of the function. As $\mu$ increases, the probability $p^*$ decreases more slowly, and vice versa. We investigate the impact of $\mu$ in Figure \ref{fig:decay}, and find that (1) the model performs worse when $p^*$ drops too quickly, and (2) when $\mu$ is within a reasonable range, there is not much impact on the final BLEU score. We use $\mu=15$ in our experiments.

\input{Figures/figure_bleu_sentence_length}

\input{Figures/figure_decay}

%% file: Figures/figure_cos_dist_cmp.tex
% \begin{wrapfigure}{r}{0.48\linewidth}
\begin{figure}
    \centering
    \includegraphics[width=\linewidth]{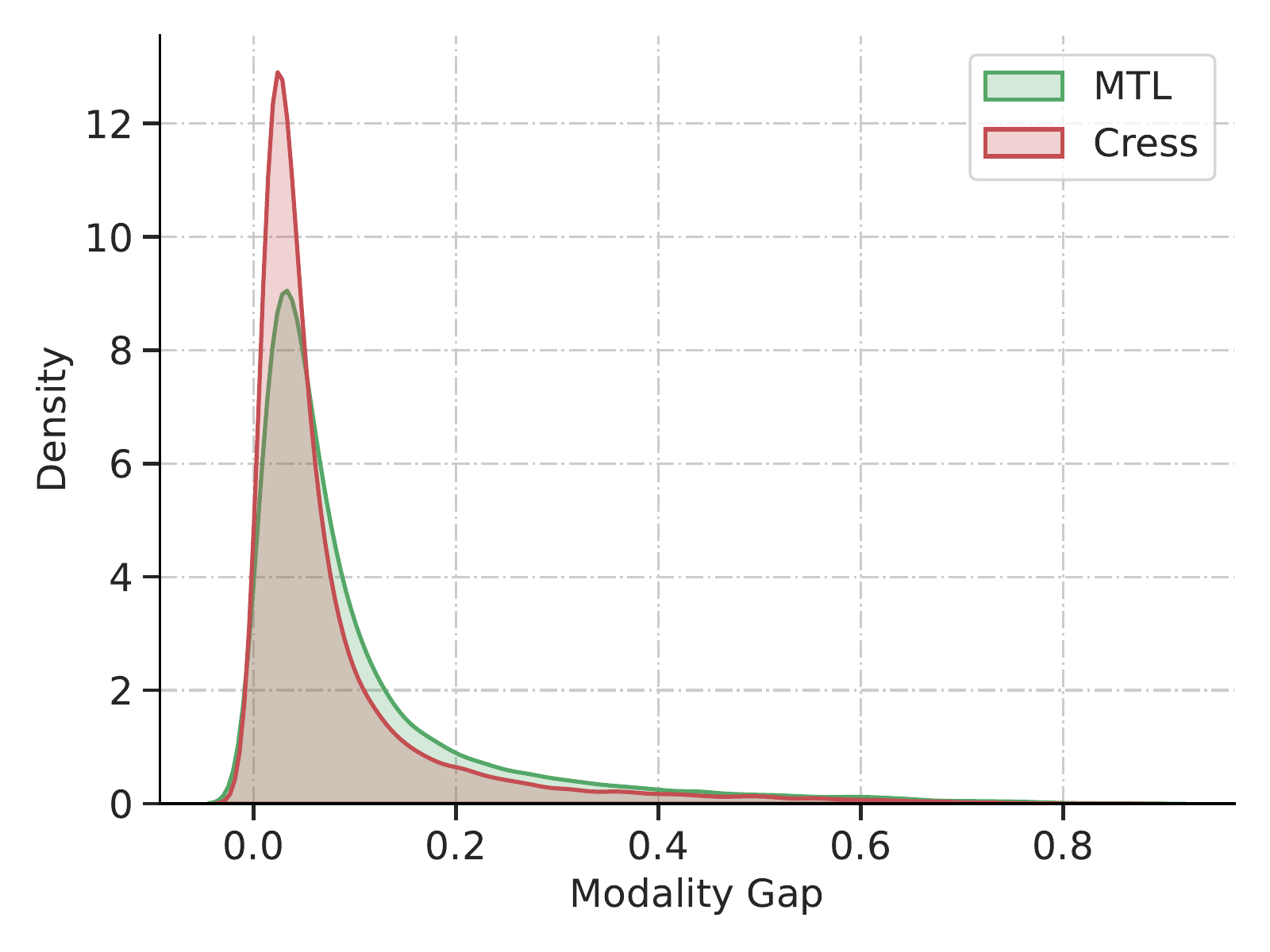}
    \caption{Distributions of the modality gap on MuST-C En$\rightarrow$De \texttt{dev} set of \textbf{\textsc{MTL}} and \textbf{\textsc{Cress}} with kernel density estimation (KDE).}
    \label{fig:cos_dist_cmp}
\end{figure}
% \end{wrapfigure}

%% file: Figures/figure_modality_gap_cmp.tex
% \begin{wrapfigure}{r}{0.45\linewidth}
\begin{figure}[t]
    \centering
    \includegraphics[width=\linewidth]{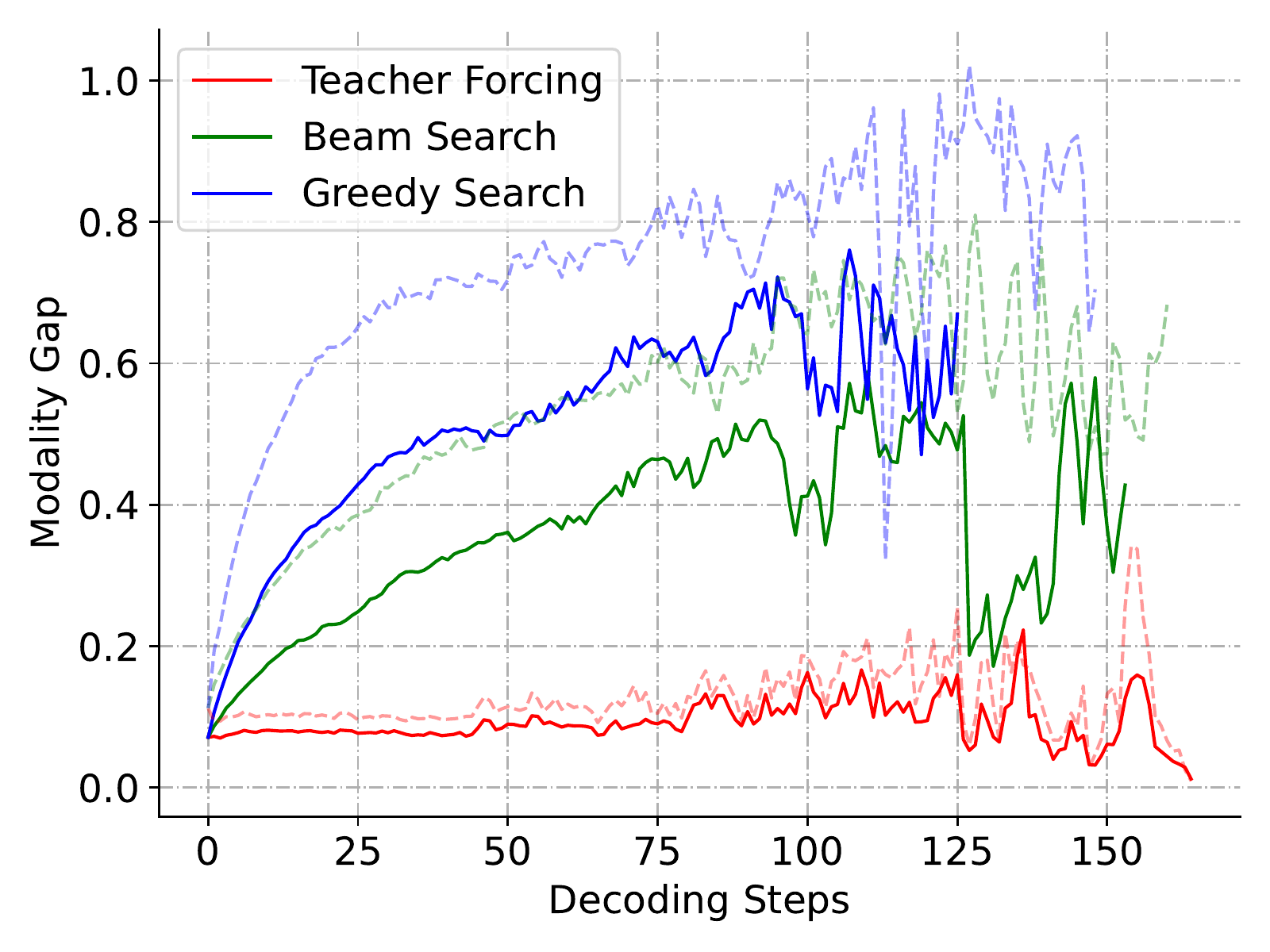}
    \caption{Curves of the average modality gap with decoding steps under three strategies. The dotted line refers to \textbf{\textsc{MTL}}, and the solid line refers to \textbf{\textsc{CRESS}}.}
    \label{fig:modality_gap_cmp}
\end{figure}
% \end{wrapfigure}

%% file: Figures/figure_heatmap.tex
% \begin{wrapfigure}{r}{0.48\linewidth}
\begin{figure}[t]
    \centering
    \includegraphics[width=\linewidth]{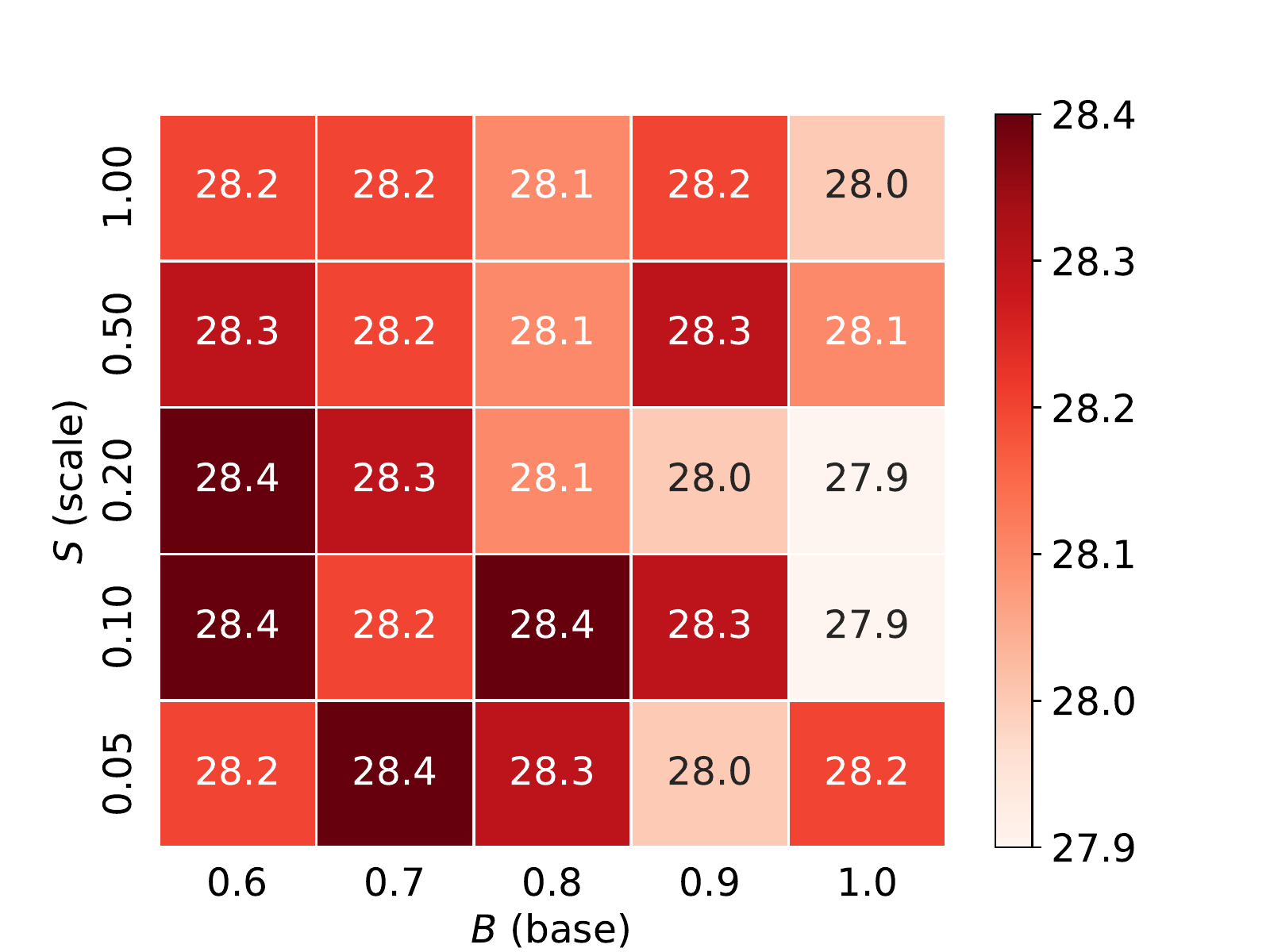}
    \caption{The heat map of BLEU scores on MuST-C En$\rightarrow$De \texttt{dev} set with different combinations of $B$ and $S$. The BLEU score without token-level adaptive training is 28.0.}
    \label{fig:heatmap}
\end{figure}
% \end{wrapfigure}

%% file: Tables/table_mt.tex
\begin{table*}[h]
\centering
\resizebox{\linewidth}{!}{
\begin{tabular}{l|c|cccccccc|cc}
\toprule
\multirow{2}{*}{\textbf{Models}} & \multirow{2}{*}{\textbf{Task}} & \multicolumn{9}{c}{\textbf{BLEU}} \\
                      &  & En$\rightarrow$De & En$\rightarrow$Fr & En$\rightarrow$Es & En$\rightarrow$Ro & En$\rightarrow$Ru & En$\rightarrow$It & En$\rightarrow$Pt & En$\rightarrow$Nl & Avg.$\uparrow$ & $\Delta\downarrow$ \\ 
\midrule
\multirow{2}{*}{\textbf{\textsc{MTL}}}  & ST  & 27.7 & 38.5 & 32.8 & 24.9 & 19.0 & 26.5 & 32.0 & 30.8 & 29.0  & \multirow{2}{*}{6.0}  \\
  & MT  & 33.5 & \textbf{46.6} & \textbf{38.3} & 30.9 & 22.1 & 33.0 & 38.6 & 36.7 & 35.0 \\
\midrule
\multirow{2}{*}{\textbf{\textsc{Cress}}}  & ST  & \textbf{29.4} & \textbf{40.1} & \textbf{33.2} & \textbf{26.4} & \textbf{19.7} & \textbf{27.6} & \textbf{33.6} & \textbf{32.3} & \textbf{30.3} & \multirow{2}{*}{\textbf{5.0}} \\
  & MT  & \textbf{34.1} & \textbf{46.6} & 38.1 & \textbf{31.1} & \textbf{22.4} & \textbf{33.3} & \textbf{39.5} & \textbf{37.6} & \textbf{35.3} \\
\bottomrule
\end{tabular}}
\caption{BLEU scores of both ST and MT on MuST-C \texttt{tst-COMMON} set (expanded setting). $\Delta$ indicates the average gap in BLEU between ST and MT.}
\label{tab:mt}
\end{table*}

%% file: Figures/figure_bleu_sentence_length.tex
% \begin{wrapfigure}{r}{0.48\linewidth}
\begin{figure}[t]
    \centering
    \includegraphics[width=\linewidth]{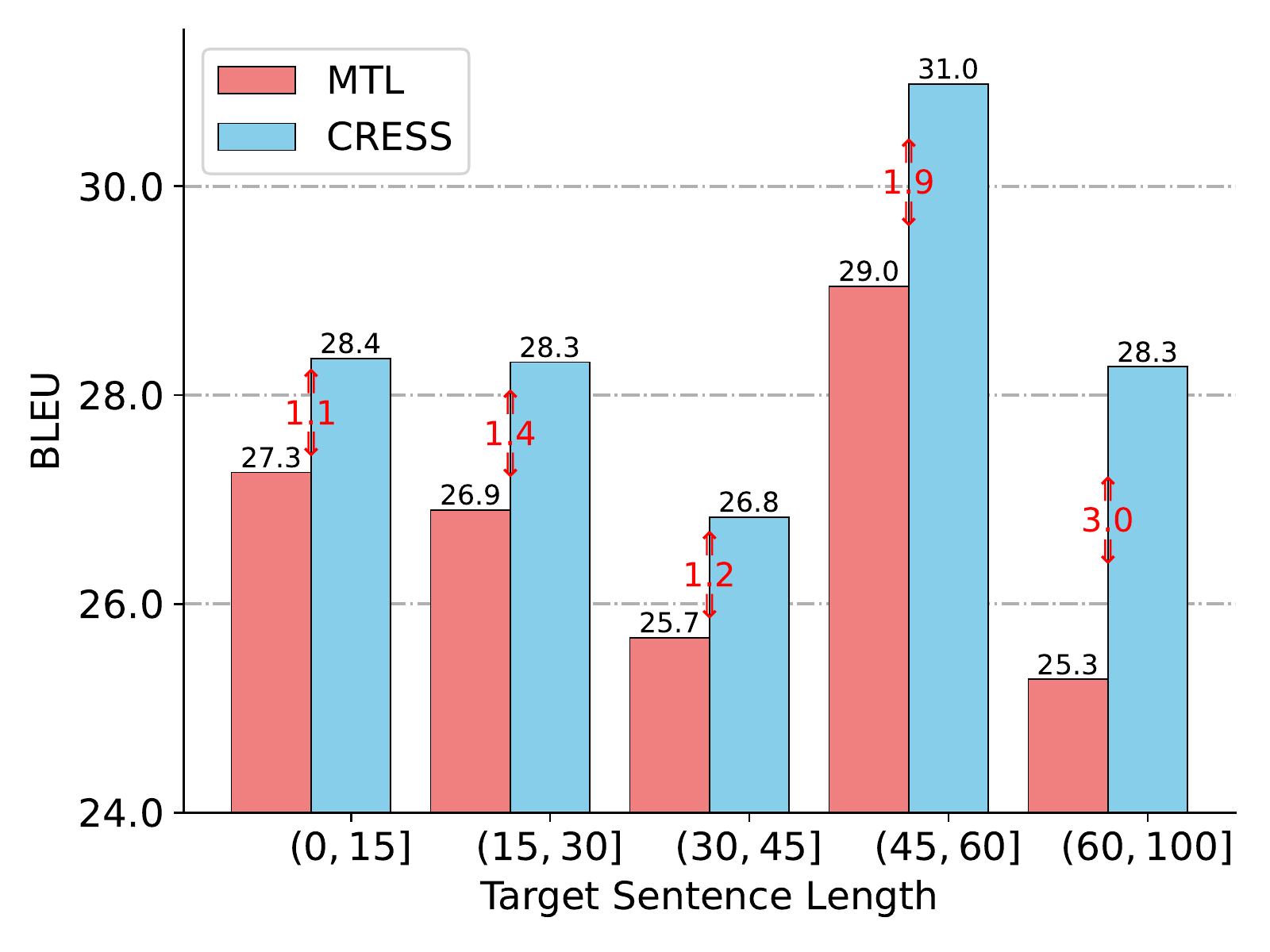}
    \caption{BLEU scores on MuST-C En$\rightarrow$De \texttt{dev} set at different target sentence lengths.}
    \label{fig:bleu_sentence_length}
\end{figure}
% \end{wrapfigure}

%% file: Figures/figure_decay.tex
\begin{figure}[t]
    \centering
    \includegraphics[width=\linewidth]{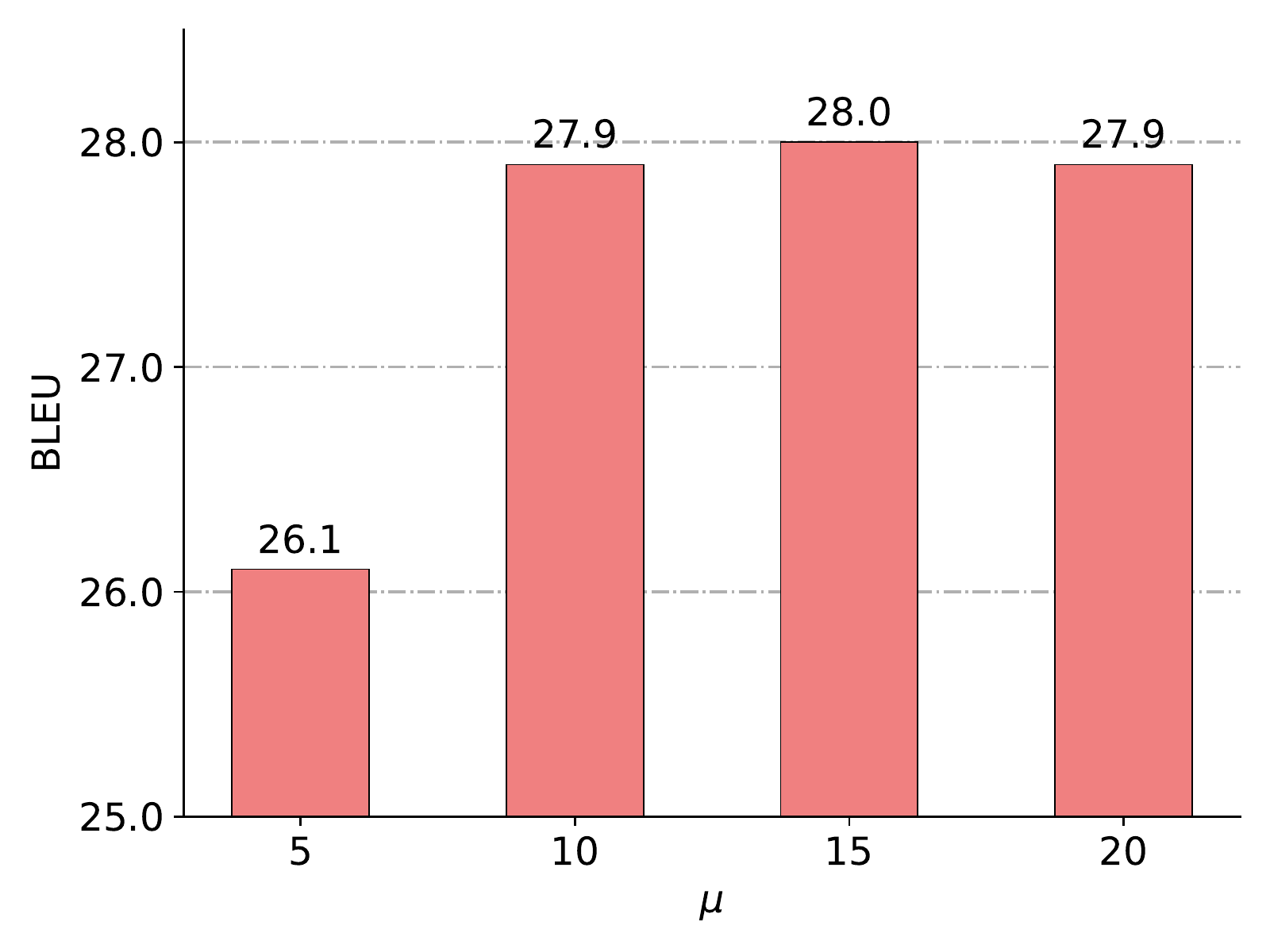}
    \caption{BLEU scores on MuST-C En$\rightarrow$De \texttt{dev} set (expanded setting) with different $\mu$. Here token-level adaptive training is not used for training.}
    \label{fig:decay}
\end{figure}

%% file: Sections/700_related_work.tex
\paragraph{End-to-end Speech Translation}
End-to-end speech translation \citep{berard2016listen, weiss2017sequence} has shown great potential for overcoming error propagation and reducing latency compared to traditional cascaded ST systems \citep{salesky2019fluent, digangi2019adapting, digangi2019enhancing, bahar2019comparative}. One challenge in training end-to-end ST models is the scarcity of ST data. To address this problem, researchers employed MT data to help training with techniques like pre-training \citep{bansal2019pre, stoian2020analyzing, wang2020bridging, wang-etal-2020-curriculum, alinejad2020effectively, le-etal-2021-lightweight, dong2021consecutive, zheng2021fused, xu-etal-2021-stacked, tang-etal-2022-unified}, multi-task learning \citep{le-etal-2020-dual, dong2021listen, ye2021end, tang-etal-2021-improving, generalmtl, taskaware}, knowledge distillation \citep{liu2019end, inaguma2021source}, and data augmentation \citep{jia2019leveraging, bahar2019using, lam-etal-2022-sample, fang-and-feng-2023-back}. However, due to the \emph{modality gap} between speech and text, it is still difficult to fully exploit MT data with the above techniques. To overcome the modality gap, \citet{han-etal-2021-learning} projects features of both speech and text into a shared semantic space. \citet{fang-etal-2022-stemm, zhou-etal-2023-cmot} mixes up features of speech and text to learn similar representations for them. \citet{ye-etal-2022-cross} brings sentence-level representations closer with contrastive learning. \citet{slam, mSLAM, Chen2022MAESTROMS, tang-etal-2022-unified, zhang2022speechut} jointly train on speech and text and design methods to align two modalities. Different from previous work, in this work, we understand the modality gap from the target-side representation differences, and show its connection to exposure bias. Based on this, we propose the \textbf{C}ross-modal \textbf{Re}gularization with \textbf{S}cheduled \textbf{S}ampling (\textbf{\textsc{Cress}}) method to bridge the modality gap.
\paragraph{Exposure Bias}
Exposure bias indicates the discrepancy between training and inference. Several approaches employ Reinforcement Learning (RL) \citep{sequence-level-training, shen-etal-2016-minimum, bahdanau2017an} instead of Maximum Likelihood Estimation (MLE) to avoid this problem. However, \citet{wu-etal-2018-study} shows that RL-based training is unstable due to the high variance of gradient estimation. An alternative and simpler approach is scheduled sampling \citep{schedule-sampling}, which samples between ground truth words and self-generated words with a changing probability. \citet{zhang-etal-2019-bridging} extends it with Gumbel noise for more robust training. In this paper, we adopt this approach to approximate the inference mode due to its training stability and low training cost.
\paragraph{Output Regularization for MT}
Regularization in the output space has proved useful for MT. \citet{rdrop} proposes to regularize the output predictions of two sub-models sampled by dropout. \citet{guo-etal-2022-prediction-difference} regularizes the output predictions of models before and after input perturbation. In this paper, we regularize the output predictions across modalities, which encourages more consistent predictions for ST and MT.
\paragraph{Token-level Adaptive Training}
Token-level adaptive training for MT is first proposed in \citet{gu-etal-2020-token}, which assigns larger weights to low-frequency words to prevent them from being ignored. \citet{xu-etal-2021-bilingual, zhang-etal-2022-conditional} computes the weight with bilingual mutual information. In this paper, we compute the weights with the modality gap between ST and MT.

%% file: Sections/800_conclusion.tex
In this paper, we propose a simple yet effective method \textbf{\textsc{Cress}} to regularize the model predictions of ST and MT, whose target-side contexts contain both ground truth words and self-generated words with scheduled sampling. Based on this, we further propose a token-level adaptive training method to handle difficult cases. Our method achieves promising results on MuST-C benchmark. Further analysis shows that our method can effectively bridge the modality gap and improve the translation quality, especially for long sentences. In the future, we will explore how to apply our method to other tasks.

%% file: Sections/X00_limitations.tex
Although our proposed method achieves promising results and outperforms most baseline systems on the ST benchmark, it still has some limitations: (1) the performance of our method still lags behind a recent work SpeechUT, although our approach has the advantage of consuming less time and resources; (2) we observe that the modality gap is still not eliminated and the effect of exposure bias on the modality gap still exists; (3) the performance of our approach on larger datasets and larger models remains to be explored; (4) how to apply our approach to other tasks also needs to be further investigated.

%% file: Sections/X00_ethics_statement.tex
In this paper, we present an effective method \textbf{\textsc{Cress}} for speech translation. While our model achieves superior performance on the widely used ST benchmark MuST-C, applying it directly to real scenarios is still risky. This is due to the fact that our training corpus only contains hundreds of hours of audio recordings from TED talks, which is far from covering all domains of the real world. Besides, the datasets we used in this paper (MuST-C, WMT, and OPUS-100) are all publicly available. We also release the code implemented with a well-known framework fairseq. These guarantee the reproducibility of our work.

%% file: Sections/X00_acknowledgement.tex
We thank all the anonymous reviewers for their insightful and valuable comments. This work was supported by National Key R\&D Program of China (NO. 2018AAA0102502)

%% file: Sections/A00_statistics.tex
\input{Tables/table_statistics}
\label{sec:statistics}

%% file: Tables/table_statistics.tex
\begin{table}[h]
    \centering
    \resizebox{\linewidth}{!}{
    \begin{tabular}{c|cr|cr}
        \toprule
         & \multicolumn{2}{c|}{\textbf{ST (MuST-C)}} & \multicolumn{2}{c}{\textbf{External MT}} \\
         \textbf{En$\rightarrow$} & hours & \#sents & name & \#sents \\
        \midrule
         \textbf{De} & 408 & 234K & WMT16 & 3.9M \\
         \textbf{Fr} & 492 & 280K & WMT14 & 31.2M \\
         \textbf{Es} & 504 & 270K & WMT13 & 14.2M \\
         \textbf{Ro} & 432 & 240K & WMT16 & 0.6M \\
         \textbf{Ru} & 489 & 270K & WMT16 & 1.9M \\
         \textbf{It} & 465 & 258K & OPUS100 & 0.7M \\
         \textbf{Pt} & 385 & 211K & OPUS100 & 0.7M \\
         \textbf{Nl} & 442 & 253K & OPUS100 & 0.7M \\
        \bottomrule
    \end{tabular}}
    \caption{Statistics of all datasets. \#sents refers to the number of parallel sentence pairs.}
    \label{tab:data}
\end{table}

%% file: Sections/C00_impact_of_acoustic_encoder.tex
\label{sec:acoustic_encoder}
Our model is composed of an acoustic encoder and a translation model. To investigate the impact of different acoustic encoders, we also conduct experiments using Wav2vec 2.0\footnote{\url{https://dl.fbaipublicfiles.com/fairseq/wav2vec/wav2vec_small.pt}} \citep{baevski2020wav2vec} as the acoustic encoder. As shown in Table \ref{tab:wav2vec}, we find that (1) HuBERT performs slightly better than Wav2vec 2.0 with an improvement of 0.5 BLEU, and (2) our proposed \textbf{\textsc{Cress}} achieves consistent improvements with different acoustic encoders. In practice, we use HuBERT to build our systems, since we believe that developing on a strong baseline will make our results more convincing and demonstrate the robustness of our approach.

\input{Tables/table_wav2vec}

%% file: Tables/table_wav2vec.tex
\begin{table}[h]
\centering
\resizebox{\linewidth}{!}{
\begin{tabular}{c|cc}
\toprule
    \textbf{Acoustic Encoder} & \textbf{\textsc{MTL}} & \textbf{\textsc{Cress}} \\
    \midrule
    HuBERT \citep{hubert}   &  27.5  &  \textbf{29.4} \\
    Wav2vec 2.0 \citep{baevski2020wav2vec} & 27.0 & \textbf{28.9} \\
\bottomrule
\end{tabular}}
\caption{BLEU scores on MuST-C En$\rightarrow$De \texttt{tst-COMMON} set (expanded setting) with different acoustic encoders.}
\label{tab:wav2vec}
\end{table}

%% file: Sections/G00_covost2.tex
\label{sec:covost}
\input{Tables/table_covost2}
We also conduct experiments on CoVoST 2 \citep{covost2} to examine the performance of our approach on large datasets. CoVoST 2 is a large-scale multilingual speech translation corpus that covers translations from 21 languages into English and from English into 15 languages. It is one of the largest open ST datasets available currently. In this paper, we evaluate our approach on the En$\rightarrow$De direction, which contains 430 hours of speech with annotated transcriptions and translations. We use \texttt{dev} set for validation and \texttt{test} set for evaluation.

We use the same pre-processing, model configuration, and hyper-parameters as MuST-C (see details in Section \ref{sec:setup}). The results are shown in Table \ref{tab:covost2}. First, we find our \textbf{\textsc{Cress}} significantly outperforms the \textbf{\textsc{MTL}} baseline, with 1.8 BLEU improvement in the base setting and 1.6 BLEU improvement in the expanded setting, which demonstrates the effectiveness and generalization capability of our method across different datasets, especially on the large-scale dataset. Second, our result is competitive with previous methods, although they use larger audio datasets ($\geq$60K hours) and larger model size ($\geq$300M), while we only use 960 hours of audio data and 155M model parameters.

%% file: Tables/table_covost2.tex
\begin{table*}[t]
\centering
\resizebox{\linewidth}{!}{
\begin{tabular}{l|c|c|c}
\toprule
\textbf{Models} & \textbf{Audio Datasets} & \textbf{\#Params} & \textbf{BLEU} \\
\midrule
wav2vec-2.0 (LS-960) \citep{DBLP:conf/interspeech/WangWPBAC21} & LS-960 & 300M & 20.5 \\
wav2vec-2.0 (LV-60K) \citep{DBLP:conf/interspeech/WangWPBAC21} & LV-60K & 300M & 25.5 \\
wav2vec-2.0 + Self-training (LV-60K) \citep{DBLP:conf/interspeech/WangWPBAC21} & LV-60K & 300M & \textbf{27.1} \\
% LNA-E,D \citep{li-etal-2021-multilingual} & LV-60K & 171M & 25.3 \\
LNA (Joint Training) \citep{li-etal-2021-multilingual} & LV-60K & 1.05B & 25.8 \\
SLAM-TLM \citep{slam} & LV-60K & 600M & \textbf{27.5} \\
XLS-R (0.3B) \citep{xls-r} & VP-400K, MLS, CV, VL, BBL & 317M & 23.6 \\
XLS-R (1B) \citep{xls-r} & VP-400K, MLS, CV, VL, BBL & 965M & 26.2 \\
XLS-R (2B) \citep{xls-r} & VP-400K, MLS, CV, VL, BBL & 2162M & \textbf{28.3} \\
\midrule
\textbf{\textsc{MTL}} (base setting) & LS-960 & \textbf{155M} & 21.4 \\
\textbf{\textsc{Cress}} (base setting) & LS-960 & \textbf{155M} & 23.2 (+1.8) \\
\textbf{\textsc{MTL}} (expanded setting) & LS-960 & \textbf{155M} & 25.1 \\
\textbf{\textsc{Cress}} (expanded setting) & LS-960 & \textbf{155M} & \textbf{26.7} (+1.6) \\
\bottomrule
\end{tabular}}
\caption{BLEU scores on CoVoST 2 En$\rightarrow$De \texttt{test} set. LS-960: LibriSpeech \citep{librispeech} (960 hours). LV-60K: Libri-Light \citep{librilight} (60K hours). VP-400K: VoxPopuli \citep{wang-etal-2021-voxpopuli} (372K hours). MLS: Multilingual LibriSpeech \citep{mlsdataset} (50K hours). CV: CommonVoice \citep{commonvoice} (7K hours). VL: VoxLingua107 \citep{voxlingua107} (6.6K hours). BBL: BABEL \citep{babel} (1K hours).}
\label{tab:covost2}
\end{table*}

%% file: Sections/B00_discussion_training_cost.tex
During training, our approach requires an additional forward pass to select predicted words compared with the baseline, which will impair the training speed. Practically, we find the training time for 1 epoch of \textbf{\textsc{Cress}} is 1.12 times longer than \textbf{\textsc{MTL}}, which is actually negligible. This is because the step of selecting predicted words is fully parallel and has no gradient calculation, which does not incur a significant time overhead.

%% file: Sections/E00_results_on_chrf.tex
\label{sec:chrf}
We also report ChrF++ score \citep{popovic-2017-chrf} using sacreBLEU toolkit\footnote{sacreBLEU signature: nrefs:1 | bs:1000 | seed:12345 | case:mixed | eff:yes | nc:6 | nw:0 | space:no | version:2.0.0} on MuST-C dataset in Table \ref{tab:chrf}. We observe that \textbf{\textsc{Cress}} outperforms \textbf{\textsc{MTL}} with 1.4 ChrF++ improvement in the base setting and 1.0 ChrF++ improvement in the expanded setting.
\input{Tables/table_chrf}

%% file: Tables/table_chrf.tex
\begin{table*}[h]
\centering
\resizebox{\linewidth}{!}{
\begin{tabular}{l|cccccccc|c}
\toprule
\multirow{2}{*}{\textbf{Models}} & \multicolumn{9}{c}{\textbf{ChrF++}} \\
                        & En$\rightarrow$De & En$\rightarrow$Fr & En$\rightarrow$Es & En$\rightarrow$Ro & En$\rightarrow$Ru & En$\rightarrow$It & En$\rightarrow$Pt & En$\rightarrow$Nl & Avg. \\ 
\midrule
\multicolumn{10}{c}{\textbf{Base setting} (w/o external MT data)} \\
\midrule
\textbf{\textsc{MTL}}                  & 52.4 & 60.4 & 56.4 & 50.9    & 41.7   & 52.6  & 57.3  & 56.1   &  53.5 \\
\textbf{\textsc{Cress}}               & \textbf{54.0**} & \textbf{62.0**} & \textbf{57.6**} & \textbf{52.4**} & \textbf{43.1**}    & \textbf{53.8**}    & \textbf{58.5**}    & \textbf{57.6**}    & \textbf{54.9}    \\
\midrule
\multicolumn{10}{c}{\textbf{Expanded setting} (w/ external MT data)} \\
\midrule
\textbf{\textsc{MTL}}                  & 54.9 & 62.6 & 58.6 & 51.9    & 44.2    & 53.4    & 57.9    & 56.9    & 55.0    \\
\textbf{\textsc{Cress}}               & \textbf{56.1**} & \textbf{63.7**} & \textbf{58.9*} & \textbf{53.1**}   & \textbf{44.5*}    & \textbf{54.2**}    & \textbf{59.3**}    & \textbf{58.3**}    &\textbf{56.0}    \\
\bottomrule
\end{tabular}}
\caption{ChrF++ scores on MuST-C \texttt{tst-COMMON} set. The external MT datasets are only used in the expanded setting. * and ** mean the improvements over \textbf{\textsc{MTL}} baseline are statistically significant ($p < 0.05$ and $p < 0.01$, respectively).}
\label{tab:chrf}
\end{table*}

%% file: acl2023.bbl
\begin{thebibliography}{78}
\expandafter\ifx\csname natexlab\endcsname\relax\def\natexlab#1{#1}\fi

\bibitem[{Alinejad and Sarkar(2020)}]{alinejad2020effectively}
Ashkan Alinejad and Anoop Sarkar. 2020.
\newblock \href {https://doi.org/10.18653/v1/2020.emnlp-main.644} {Effectively
  pretraining a speech translation decoder with machine translation data}.
\newblock In \emph{Proceedings of the 2020 Conference on Empirical Methods in
  Natural Language Processing (EMNLP)}, pages 8014--8020, Online. Association
  for Computational Linguistics.

\bibitem[{Anastasopoulos et~al.(2022)Anastasopoulos, Barrault, Bentivogli,
  Zanon~Boito, Bojar, Cattoni, Currey, Dinu, Duh, Elbayad, Emmanuel,
  Est{\`e}ve, Federico, Federmann, Gahbiche, Gong, Grundkiewicz, Haddow, Hsu,
  Javorsk{\'y}, Kloudov{\'a}, Lakew, Ma, Mathur, McNamee, Murray,
  N{\v{a}}dejde, Nakamura, Negri, Niehues, Niu, Ortega, Pino, Salesky, Shi,
  Sperber, St{\"u}ker, Sudoh, Turchi, Virkar, Waibel, Wang, and
  Watanabe}]{anastasopoulos-etal-2022-findings}
Antonios Anastasopoulos, Lo{\"\i}c Barrault, Luisa Bentivogli, Marcely
  Zanon~Boito, Ond{\v{r}}ej Bojar, Roldano Cattoni, Anna Currey, Georgiana
  Dinu, Kevin Duh, Maha Elbayad, Clara Emmanuel, Yannick Est{\`e}ve, Marcello
  Federico, Christian Federmann, Souhir Gahbiche, Hongyu Gong, Roman
  Grundkiewicz, Barry Haddow, Benjamin Hsu, D{\'a}vid Javorsk{\'y}, V{\u{e}}ra
  Kloudov{\'a}, Surafel Lakew, Xutai Ma, Prashant Mathur, Paul McNamee, Kenton
  Murray, Maria N{\v{a}}dejde, Satoshi Nakamura, Matteo Negri, Jan Niehues,
  Xing Niu, John Ortega, Juan Pino, Elizabeth Salesky, Jiatong Shi, Matthias
  Sperber, Sebastian St{\"u}ker, Katsuhito Sudoh, Marco Turchi, Yogesh Virkar,
  Alexander Waibel, Changhan Wang, and Shinji Watanabe. 2022.
\newblock \href {https://doi.org/10.18653/v1/2022.iwslt-1.10} {Findings of the
  {IWSLT} 2022 evaluation campaign}.
\newblock In \emph{Proceedings of the 19th International Conference on Spoken
  Language Translation (IWSLT 2022)}, pages 98--157, Dublin, Ireland (in-person
  and online). Association for Computational Linguistics.

\bibitem[{Anastasopoulos et~al.(2021)Anastasopoulos, Bojar, Bremerman, Cattoni,
  Elbayad, Federico, Ma, Nakamura, Negri, Niehues, Pino, Salesky, St{\"u}ker,
  Sudoh, Turchi, Waibel, Wang, and Wiesner}]{anastasopoulos-etal-2021-findings}
Antonios Anastasopoulos, Ond{\v{r}}ej Bojar, Jacob Bremerman, Roldano Cattoni,
  Maha Elbayad, Marcello Federico, Xutai Ma, Satoshi Nakamura, Matteo Negri,
  Jan Niehues, Juan Pino, Elizabeth Salesky, Sebastian St{\"u}ker, Katsuhito
  Sudoh, Marco Turchi, Alexander Waibel, Changhan Wang, and Matthew Wiesner.
  2021.
\newblock \href {https://doi.org/10.18653/v1/2021.iwslt-1.1} {{FINDINGS} {OF}
  {THE} {IWSLT} 2021 {EVALUATION} {CAMPAIGN}}.
\newblock In \emph{Proceedings of the 18th International Conference on Spoken
  Language Translation (IWSLT 2021)}, pages 1--29, Bangkok, Thailand (online).
  Association for Computational Linguistics.

\bibitem[{Ardila et~al.(2020)Ardila, Branson, Davis, Kohler, Meyer, Henretty,
  Morais, Saunders, Tyers, and Weber}]{commonvoice}
Rosana Ardila, Megan Branson, Kelly Davis, Michael Kohler, Josh Meyer, Michael
  Henretty, Reuben Morais, Lindsay Saunders, Francis~M. Tyers, and Gregor
  Weber. 2020.
\newblock \href {https://aclanthology.org/2020.lrec-1.520/} {Common voice: {A}
  massively-multilingual speech corpus}.
\newblock In \emph{Proceedings of The 12th Language Resources and Evaluation
  Conference, {LREC} 2020, Marseille, France, May 11-16, 2020}, pages
  4218--4222. European Language Resources Association.

\bibitem[{Arora et~al.(2022)Arora, El~Asri, Bahuleyan, and
  Cheung}]{arora-etal-2022-exposure}
Kushal Arora, Layla El~Asri, Hareesh Bahuleyan, and Jackie Cheung. 2022.
\newblock \href {https://doi.org/10.18653/v1/2022.findings-acl.58} {Why
  exposure bias matters: An imitation learning perspective of error
  accumulation in language generation}.
\newblock In \emph{Findings of the Association for Computational Linguistics:
  ACL 2022}, pages 700--710, Dublin, Ireland. Association for Computational
  Linguistics.

\bibitem[{Babu et~al.(2022)Babu, Wang, Tjandra, Lakhotia, Xu, Goyal, Singh, von
  Platen, Saraf, Pino, Baevski, Conneau, and Auli}]{xls-r}
Arun Babu, Changhan Wang, Andros Tjandra, Kushal Lakhotia, Qiantong Xu, Naman
  Goyal, Kritika Singh, Patrick von Platen, Yatharth Saraf, Juan Pino, Alexei
  Baevski, Alexis Conneau, and Michael Auli. 2022.
\newblock \href {https://doi.org/10.21437/Interspeech.2022-143} {{XLS-R:}
  self-supervised cross-lingual speech representation learning at scale}.
\newblock In \emph{Interspeech 2022, 23rd Annual Conference of the
  International Speech Communication Association, Incheon, Korea, 18-22
  September 2022}, pages 2278--2282. {ISCA}.

\bibitem[{Baevski et~al.(2020)Baevski, Zhou, Mohamed, and
  Auli}]{baevski2020wav2vec}
Alexei Baevski, Yuhao Zhou, Abdelrahman Mohamed, and Michael Auli. 2020.
\newblock wav2vec 2.0: A framework for self-supervised learning of speech
  representations.
\newblock \emph{Advances in Neural Information Processing Systems}, 33.

\bibitem[{Bahar et~al.(2019{\natexlab{a}})Bahar, Bieschke, and
  Ney}]{bahar2019comparative}
Parnia Bahar, Tobias Bieschke, and Hermann Ney. 2019{\natexlab{a}}.
\newblock A comparative study on end-to-end speech to text translation.
\newblock In \emph{Proc. of ASRU}, pages 792--799. IEEE.

\bibitem[{Bahar et~al.(2019{\natexlab{b}})Bahar, Zeyer, Schl{\"u}ter, and
  Ney}]{bahar2019using}
Parnia Bahar, Albert Zeyer, Ralf Schl{\"u}ter, and Hermann Ney.
  2019{\natexlab{b}}.
\newblock On using specaugment for end-to-end speech translation.
\newblock In \emph{Proc. of IWSLT}.

\bibitem[{Bahdanau et~al.(2017)Bahdanau, Brakel, Xu, Goyal, Lowe, Pineau,
  Courville, and Bengio}]{bahdanau2017an}
Dzmitry Bahdanau, Philemon Brakel, Kelvin Xu, Anirudh Goyal, Ryan Lowe, Joelle
  Pineau, Aaron Courville, and Yoshua Bengio. 2017.
\newblock \href {https://openreview.net/forum?id=SJDaqqveg} {An actor-critic
  algorithm for sequence prediction}.
\newblock In \emph{International Conference on Learning Representations}.

\bibitem[{Bansal et~al.(2019)Bansal, Kamper, Livescu, Lopez, and
  Goldwater}]{bansal2019pre}
Sameer Bansal, Herman Kamper, Karen Livescu, Adam Lopez, and Sharon Goldwater.
  2019.
\newblock Pre-training on high-resource speech recognition improves
  low-resource speech-to-text translation.
\newblock In \emph{Proc. of NAACL-HLT}, pages 58--68.

\bibitem[{Bapna et~al.(2022)Bapna, Cherry, Zhang, Jia, Johnson, Cheng, Khanuja,
  Riesa, and Conneau}]{mSLAM}
Ankur Bapna, Colin Cherry, Yu~Zhang, Ye~Jia, Melvin Johnson, Yong Cheng, Simran
  Khanuja, Jason Riesa, and Alexis Conneau. 2022.
\newblock \href {http://arxiv.org/abs/2202.01374} {mslam: Massively
  multilingual joint pre-training for speech and text}.
\newblock \emph{CoRR}, abs/2202.01374.

\bibitem[{Bapna et~al.(2021)Bapna, Chung, Wu, Gulati, Jia, Clark, Johnson,
  Riesa, Conneau, and Zhang}]{slam}
Ankur Bapna, Yu-an Chung, Nan Wu, Anmol Gulati, Ye~Jia, Jonathan~H Clark,
  Melvin Johnson, Jason Riesa, Alexis Conneau, and Yu~Zhang. 2021.
\newblock Slam: A unified encoder for speech and language modeling via
  speech-text joint pre-training.
\newblock \emph{arXiv preprint arXiv:2110.10329}.

\bibitem[{Bengio et~al.(2015)Bengio, Vinyals, Jaitly, and
  Shazeer}]{schedule-sampling}
Samy Bengio, Oriol Vinyals, Navdeep Jaitly, and Noam Shazeer. 2015.
\newblock \href
  {https://proceedings.neurips.cc/paper/2015/file/e995f98d56967d946471af29d7bf99f1-Paper.pdf}
  {Scheduled sampling for sequence prediction with recurrent neural networks}.
\newblock In \emph{Advances in Neural Information Processing Systems},
  volume~28. Curran Associates, Inc.

\bibitem[{Bentivogli et~al.(2021)Bentivogli, Cettolo, Gaido, Karakanta,
  Martinelli, Negri, and Turchi}]{bentivogli-etal-2021-cascade}
Luisa Bentivogli, Mauro Cettolo, Marco Gaido, Alina Karakanta, Alberto
  Martinelli, Matteo Negri, and Marco Turchi. 2021.
\newblock \href {https://doi.org/10.18653/v1/2021.acl-long.224} {Cascade versus
  direct speech translation: Do the differences still make a difference?}
\newblock In \emph{Proceedings of the 59th Annual Meeting of the Association
  for Computational Linguistics and the 11th International Joint Conference on
  Natural Language Processing (Volume 1: Long Papers)}, pages 2873--2887,
  Online. Association for Computational Linguistics.

\bibitem[{B{\'e}rard et~al.(2016)B{\'e}rard, Pietquin, Servan, and
  Besacier}]{berard2016listen}
Alexandre B{\'e}rard, Olivier Pietquin, Christophe Servan, and Laurent
  Besacier. 2016.
\newblock Listen and translate: A proof of concept for end-to-end
  speech-to-text translation.
\newblock In \emph{NIPS workshop on End-to-end Learning for Speech and Audio
  Processing}.

\bibitem[{Buck and Koehn(2016)}]{buck-koehn-2016-findings}
Christian Buck and Philipp Koehn. 2016.
\newblock \href {https://doi.org/10.18653/v1/W16-2347} {Findings of the {WMT}
  2016 bilingual document alignment shared task}.
\newblock In \emph{Proceedings of the First Conference on Machine Translation:
  Volume 2, Shared Task Papers}, pages 554--563, Berlin, Germany. Association
  for Computational Linguistics.

\bibitem[{Chen et~al.(2022)Chen, Zhang, Rosenberg, Ramabhadran, Moreno, Bapna,
  and Zen}]{Chen2022MAESTROMS}
Zhehuai Chen, Yu~Zhang, Andrew Rosenberg, Bhuvana Ramabhadran, Pedro~J. Moreno,
  Ankur Bapna, and Heiga Zen. 2022.
\newblock Maestro: Matched speech text representations through modality
  matching.
\newblock In \emph{INTERSPEECH}.

\bibitem[{Di~Gangi et~al.(2019{\natexlab{a}})Di~Gangi, Cattoni, Bentivogli,
  Negri, and Turchi}]{di-gangi-etal-2019-must}
Mattia~A. Di~Gangi, Roldano Cattoni, Luisa Bentivogli, Matteo Negri, and Marco
  Turchi. 2019{\natexlab{a}}.
\newblock \href {https://doi.org/10.18653/v1/N19-1202} {{M}u{ST}-{C}: a
  {M}ultilingual {S}peech {T}ranslation {C}orpus}.
\newblock In \emph{Proceedings of the 2019 Conference of the North {A}merican
  Chapter of the Association for Computational Linguistics: Human Language
  Technologies, Volume 1 (Long and Short Papers)}, pages 2012--2017,
  Minneapolis, Minnesota. Association for Computational Linguistics.

\bibitem[{Di~Gangi et~al.(2019{\natexlab{b}})Di~Gangi, Negri, Cattoni, Dessi,
  and Turchi}]{digangi2019enhancing}
Mattia~A. Di~Gangi, Matteo Negri, Roldano Cattoni, Roberto Dessi, and Marco
  Turchi. 2019{\natexlab{b}}.
\newblock Enhancing transformer for end-to-end speech-to-text translation.
\newblock In \emph{Proceedings of Machine Translation Summit XVII Volume 1:
  Research Track}, pages 21--31.

\bibitem[{Di~Gangi et~al.(2019{\natexlab{c}})Di~Gangi, Negri, and
  Turchi}]{digangi2019adapting}
Mattia~A. Di~Gangi, Matteo Negri, and Marco Turchi. 2019{\natexlab{c}}.
\newblock Adapting transformer to end-to-end spoken language translation.
\newblock In \emph{Proc. of INTERSPEECH}, pages 1133--1137. International
  Speech Communication Association (ISCA).

\bibitem[{Dong et~al.(2021{\natexlab{a}})Dong, Wang, Zhou, Xu, Xu, and
  Li}]{dong2021consecutive}
Qianqian Dong, Mingxuan Wang, Hao Zhou, Shuang Xu, Bo~Xu, and Lei Li.
  2021{\natexlab{a}}.
\newblock Consecutive decoding for speech-to-text translation.
\newblock In \emph{The Thirty-fifth AAAI Conference on Artificial Intelligence,
  AAAI}.

\bibitem[{Dong et~al.(2021{\natexlab{b}})Dong, Ye, Wang, Zhou, Xu, Xu, and
  Li}]{dong2021listen}
Qianqian Dong, Rong Ye, Mingxuan Wang, Hao Zhou, Shuang Xu, Bo~Xu, and Lei Li.
  2021{\natexlab{b}}.
\newblock Listen, understand and translate: Triple supervision decouples
  end-to-end speech-to-text translation.
\newblock In \emph{Proceedings of the AAAI Conference on Artificial
  Intelligence}.

\bibitem[{Fang and Feng(2023)}]{fang-and-feng-2023-back}
Qingkai Fang and Yang Feng. 2023.
\newblock Back translation for speech-to-text translation without transcripts.
\newblock In \emph{Proceedings of the 61st Annual Meeting of the Association
  for Computational Linguistics}.

\bibitem[{Fang et~al.(2022)Fang, Ye, Li, Feng, and Wang}]{fang-etal-2022-stemm}
Qingkai Fang, Rong Ye, Lei Li, Yang Feng, and Mingxuan Wang. 2022.
\newblock \href {https://doi.org/10.18653/v1/2022.acl-long.486} {{STEMM}:
  Self-learning with speech-text manifold mixup for speech translation}.
\newblock In \emph{Proceedings of the 60th Annual Meeting of the Association
  for Computational Linguistics (Volume 1: Long Papers)}, pages 7050--7062,
  Dublin, Ireland. Association for Computational Linguistics.

\bibitem[{Gales et~al.(2014)Gales, Knill, Ragni, and Rath}]{babel}
Mark J.~F. Gales, Kate~M. Knill, Anton Ragni, and Shakti~P. Rath. 2014.
\newblock \href
  {http://www.isca-speech.org/archive/sltu\_2014/gales14\_sltu.html} {Speech
  recognition and keyword spotting for low-resource languages: Babel project
  research at {CUED}}.
\newblock In \emph{4th Workshop on Spoken Language Technologies for
  Under-resourced Languages, {SLTU} 2014, St. Petersburg, Russia, May 14-16,
  2014}, pages 16--23. {ISCA}.

\bibitem[{Gu et~al.(2020)Gu, Zhang, Meng, Feng, Xie, Zhou, and
  Yu}]{gu-etal-2020-token}
Shuhao Gu, Jinchao Zhang, Fandong Meng, Yang Feng, Wanying Xie, Jie Zhou, and
  Dong Yu. 2020.
\newblock \href {https://doi.org/10.18653/v1/2020.emnlp-main.76} {Token-level
  adaptive training for neural machine translation}.
\newblock In \emph{Proceedings of the 2020 Conference on Empirical Methods in
  Natural Language Processing (EMNLP)}, pages 1035--1046, Online. Association
  for Computational Linguistics.

\bibitem[{Gumbel(1954)}]{gumbel1954statistical}
Emil~Julius Gumbel. 1954.
\newblock \href
  {https://ntrl.ntis.gov/NTRL/dashboard/searchResults/titleDetail/PB175818.xhtml#}
  {Statistical theory of extreme values and some practical applications: a
  series of lectures}.
\newblock In \emph{Nat. Bur. Standards Appl. Math. Ser.}, volume~33. US
  Government Printing Office.

\bibitem[{Guo et~al.(2022)Guo, Ma, Zhang, and
  Feng}]{guo-etal-2022-prediction-difference}
Dengji Guo, Zhengrui Ma, Min Zhang, and Yang Feng. 2022.
\newblock Prediction difference regularization against perturbation for neural
  machine translation.
\newblock In \emph{Proceedings of the 60th Annual Meeting of the Association
  for Computational Linguistics}.

\bibitem[{Han et~al.(2021)Han, Wang, Ji, and Li}]{han-etal-2021-learning}
Chi Han, Mingxuan Wang, Heng Ji, and Lei Li. 2021.
\newblock \href {https://doi.org/10.18653/v1/2021.findings-acl.195} {Learning
  shared semantic space for speech-to-text translation}.
\newblock In \emph{Findings of the Association for Computational Linguistics:
  ACL-IJCNLP 2021}, pages 2214--2225, Online. Association for Computational
  Linguistics.

\bibitem[{Hsu et~al.(2021)Hsu, Bolte, Tsai, Lakhotia, Salakhutdinov, and
  Mohamed}]{hubert}
Wei-Ning Hsu, Benjamin Bolte, Yao-Hung~Hubert Tsai, Kushal Lakhotia, Ruslan
  Salakhutdinov, and Abdelrahman Mohamed. 2021.
\newblock \href {https://doi.org/10.1109/TASLP.2021.3122291} {Hubert:
  Self-supervised speech representation learning by masked prediction of hidden
  units}.
\newblock \emph{IEEE/ACM Trans. Audio, Speech and Lang. Proc.}, 29:3451–3460.

\bibitem[{Inaguma et~al.(2021)Inaguma, Kawahara, and
  Watanabe}]{inaguma2021source}
Hirofumi Inaguma, Tatsuya Kawahara, and Shinji Watanabe. 2021.
\newblock Source and target bidirectional knowledge distillation for end-to-end
  speech translation.
\newblock In \emph{Proceedings of NAACL}, pages 1872--1881.

\bibitem[{Indurthi et~al.(2021)Indurthi, Zaidi, Kumar~Lakumarapu, Lee, Han,
  Ahn, Kim, Kim, and Hwang}]{taskaware}
Sathish Indurthi, Mohd~Abbas Zaidi, Nikhil Kumar~Lakumarapu, Beomseok Lee,
  Hyojung Han, Seokchan Ahn, Sangha Kim, Chanwoo Kim, and Inchul Hwang. 2021.
\newblock \href {https://doi.org/10.1109/ICASSP39728.2021.9414703} {Task aware
  multi-task learning for speech to text tasks}.
\newblock In \emph{ICASSP 2021 - 2021 IEEE International Conference on
  Acoustics, Speech and Signal Processing (ICASSP)}, pages 7723--7727.

\bibitem[{Jia et~al.(2019)Jia, Johnson, Macherey, Weiss, Cao, Chiu, Ari,
  Laurenzo, and Wu}]{jia2019leveraging}
Ye~Jia, Melvin Johnson, Wolfgang Macherey, Ron~J. Weiss, Yuan Cao,
  Chung{-}Cheng Chiu, Naveen Ari, Stella Laurenzo, and Yonghui Wu. 2019.
\newblock Leveraging weakly supervised data to improve end-to-end
  speech-to-text translation.
\newblock In \emph{Proc. of ICASSP}, pages 7180--7184.

\bibitem[{Kahn et~al.(2020)Kahn, Rivi{\`{e}}re, Zheng, Kharitonov, Xu,
  Mazar{\'{e}}, Karadayi, Liptchinsky, Collobert, Fuegen, Likhomanenko,
  Synnaeve, Joulin, Mohamed, and Dupoux}]{librilight}
Jacob Kahn, Morgane Rivi{\`{e}}re, Weiyi Zheng, Evgeny Kharitonov, Qiantong Xu,
  Pierre{-}Emmanuel Mazar{\'{e}}, Julien Karadayi, Vitaliy Liptchinsky, Ronan
  Collobert, Christian Fuegen, Tatiana Likhomanenko, Gabriel Synnaeve, Armand
  Joulin, Abdelrahman Mohamed, and Emmanuel Dupoux. 2020.
\newblock \href {https://doi.org/10.1109/ICASSP40776.2020.9052942}
  {Libri-light: {A} benchmark for {ASR} with limited or no supervision}.
\newblock In \emph{2020 {IEEE} International Conference on Acoustics, Speech
  and Signal Processing, {ICASSP} 2020, Barcelona, Spain, May 4-8, 2020}, pages
  7669--7673. {IEEE}.

\bibitem[{Kingma and Ba(2015)}]{adam}
Diederik~P. Kingma and Jimmy Ba. 2015.
\newblock \href {http://arxiv.org/abs/1412.6980} {Adam: A method for stochastic
  optimization}.
\newblock In \emph{ICLR (Poster)}.

\bibitem[{Koehn(2004)}]{koehn-2004-statistical}
Philipp Koehn. 2004.
\newblock \href {https://aclanthology.org/W04-3250} {Statistical significance
  tests for machine translation evaluation}.
\newblock In \emph{Proceedings of the 2004 Conference on Empirical Methods in
  Natural Language Processing}, pages 388--395, Barcelona, Spain. Association
  for Computational Linguistics.

\bibitem[{Lam et~al.(2022)Lam, Schamoni, and Riezler}]{lam-etal-2022-sample}
Tsz~Kin Lam, Shigehiko Schamoni, and Stefan Riezler. 2022.
\newblock \href {https://doi.org/10.18653/v1/2022.acl-short.27} {Sample,
  translate, recombine: Leveraging audio alignments for data augmentation in
  end-to-end speech translation}.
\newblock In \emph{Proceedings of the 60th Annual Meeting of the Association
  for Computational Linguistics (Volume 2: Short Papers)}, pages 245--254,
  Dublin, Ireland. Association for Computational Linguistics.

\bibitem[{Le et~al.(2020)Le, Pino, Wang, Gu, Schwab, and
  Besacier}]{le-etal-2020-dual}
Hang Le, Juan Pino, Changhan Wang, Jiatao Gu, Didier Schwab, and Laurent
  Besacier. 2020.
\newblock \href {https://doi.org/10.18653/v1/2020.coling-main.314}
  {Dual-decoder transformer for joint automatic speech recognition and
  multilingual speech translation}.
\newblock In \emph{Proceedings of the 28th International Conference on
  Computational Linguistics}, pages 3520--3533, Barcelona, Spain (Online).
  International Committee on Computational Linguistics.

\bibitem[{Le et~al.(2021)Le, Pino, Wang, Gu, Schwab, and
  Besacier}]{le-etal-2021-lightweight}
Hang Le, Juan Pino, Changhan Wang, Jiatao Gu, Didier Schwab, and Laurent
  Besacier. 2021.
\newblock \href {https://doi.org/10.18653/v1/2021.acl-short.103} {Lightweight
  adapter tuning for multilingual speech translation}.
\newblock In \emph{Proceedings of the 59th Annual Meeting of the Association
  for Computational Linguistics and the 11th International Joint Conference on
  Natural Language Processing (Volume 2: Short Papers)}, pages 817--824,
  Online. Association for Computational Linguistics.

\bibitem[{Li et~al.(2021)Li, Wang, Tang, Tran, Tang, Pino, Baevski, Conneau,
  and Auli}]{li-etal-2021-multilingual}
Xian Li, Changhan Wang, Yun Tang, Chau Tran, Yuqing Tang, Juan Pino, Alexei
  Baevski, Alexis Conneau, and Michael Auli. 2021.
\newblock \href {https://doi.org/10.18653/v1/2021.acl-long.68} {Multilingual
  speech translation from efficient finetuning of pretrained models}.
\newblock In \emph{Proceedings of the 59th Annual Meeting of the Association
  for Computational Linguistics and the 11th International Joint Conference on
  Natural Language Processing (Volume 1: Long Papers)}, pages 827--838, Online.
  Association for Computational Linguistics.

\bibitem[{Liang et~al.(2021)Liang, Wu, Li, Wang, Meng, Qin, Chen, Zhang, and
  Liu}]{rdrop}
Xiaobo Liang, Lijun Wu, Juntao Li, Yue Wang, Qi~Meng, Tao Qin, Wei Chen, Min
  Zhang, and Tie-Yan Liu. 2021.
\newblock \href
  {https://proceedings.neurips.cc/paper/2021/file/5a66b9200f29ac3fa0ae244cc2a51b39-Paper.pdf}
  {R-drop: Regularized dropout for neural networks}.
\newblock In \emph{Advances in Neural Information Processing Systems},
  volume~34, pages 10890--10905. Curran Associates, Inc.

\bibitem[{Liu et~al.(2019)Liu, Xiong, Zhang, He, Wu, Wang, and
  Zong}]{liu2019end}
Yuchen Liu, Hao Xiong, Jiajun Zhang, Zhongjun He, Hua Wu, Haifeng Wang, and
  Chengqing Zong. 2019.
\newblock \href {https://doi.org/10.21437/Interspeech.2019-2582} {{End-to-End
  Speech Translation with Knowledge Distillation}}.
\newblock In \emph{Proc. Interspeech 2019}, pages 1128--1132.

\bibitem[{Maddison et~al.(2014)Maddison, Tarlow, and Minka}]{asampling}
Chris~J. Maddison, Daniel Tarlow, and Tom Minka. 2014.
\newblock \href
  {https://proceedings.neurips.cc/paper/2014/file/309fee4e541e51de2e41f21bebb342aa-Paper.pdf}
  {A$\ast$ sampling}.
\newblock In \emph{Advances in Neural Information Processing Systems},
  volume~27. Curran Associates, Inc.

\bibitem[{Ott et~al.(2019)Ott, Edunov, Baevski, Fan, Gross, Ng, Grangier, and
  Auli}]{ott2019fairseq}
Myle Ott, Sergey Edunov, Alexei Baevski, Angela Fan, Sam Gross, Nathan Ng,
  David Grangier, and Michael Auli. 2019.
\newblock fairseq: A fast, extensible toolkit for sequence modeling.
\newblock In \emph{Proceedings of NAACL-HLT 2019: Demonstrations}.

\bibitem[{Panayotov et~al.(2015)Panayotov, Chen, Povey, and
  Khudanpur}]{librispeech}
Vassil Panayotov, Guoguo Chen, Daniel Povey, and Sanjeev Khudanpur. 2015.
\newblock \href {https://doi.org/10.1109/ICASSP.2015.7178964} {Librispeech: An
  asr corpus based on public domain audio books}.
\newblock In \emph{2015 IEEE International Conference on Acoustics, Speech and
  Signal Processing (ICASSP)}, pages 5206--5210.

\bibitem[{Papineni et~al.(2002)Papineni, Roukos, Ward, and
  Zhu}]{papineni-etal-2002-bleu}
Kishore Papineni, Salim Roukos, Todd Ward, and Wei-Jing Zhu. 2002.
\newblock \href {https://doi.org/10.3115/1073083.1073135} {{B}leu: a method for
  automatic evaluation of machine translation}.
\newblock In \emph{Proceedings of the 40th Annual Meeting of the Association
  for Computational Linguistics}, pages 311--318, Philadelphia, Pennsylvania,
  USA. Association for Computational Linguistics.

\bibitem[{Popovi{\'c}(2017)}]{popovic-2017-chrf}
Maja Popovi{\'c}. 2017.
\newblock \href {https://doi.org/10.18653/v1/W17-4770} {chr{F}++: words helping
  character n-grams}.
\newblock In \emph{Proceedings of the Second Conference on Machine
  Translation}, pages 612--618, Copenhagen, Denmark. Association for
  Computational Linguistics.

\bibitem[{Post(2018)}]{post-2018-call}
Matt Post. 2018.
\newblock \href {https://doi.org/10.18653/v1/W18-6319} {A call for clarity in
  reporting {BLEU} scores}.
\newblock In \emph{Proceedings of the Third Conference on Machine Translation:
  Research Papers}, pages 186--191, Brussels, Belgium. Association for
  Computational Linguistics.

\bibitem[{Pratap et~al.(2020)Pratap, Xu, Sriram, Synnaeve, and
  Collobert}]{mlsdataset}
Vineel Pratap, Qiantong Xu, Anuroop Sriram, Gabriel Synnaeve, and Ronan
  Collobert. 2020.
\newblock \href {https://doi.org/10.21437/Interspeech.2020-2826} {{MLS:} {A}
  large-scale multilingual dataset for speech research}.
\newblock In \emph{Interspeech 2020, 21st Annual Conference of the
  International Speech Communication Association, Virtual Event, Shanghai,
  China, 25-29 October 2020}, pages 2757--2761. {ISCA}.

\bibitem[{Ranzato et~al.(2016)Ranzato, Chopra, Auli, and
  Zaremba}]{sequence-level-training}
Marc'Aurelio Ranzato, Sumit Chopra, Michael Auli, and Wojciech Zaremba. 2016.
\newblock \href {http://arxiv.org/abs/1511.06732} {Sequence level training with
  recurrent neural networks}.
\newblock In \emph{4th International Conference on Learning Representations,
  {ICLR} 2016, San Juan, Puerto Rico, May 2-4, 2016, Conference Track
  Proceedings}.

\bibitem[{Salesky et~al.(2019)Salesky, Sperber, and Waibel}]{salesky2019fluent}
Elizabeth Salesky, Matthias Sperber, and Alexander Waibel. 2019.
\newblock Fluent translations from disfluent speech in end-to-end speech
  translation.
\newblock In \emph{Proc. of NAACL-HLT}, pages 2786--2792.

\bibitem[{Shen et~al.(2016)Shen, Cheng, He, He, Wu, Sun, and
  Liu}]{shen-etal-2016-minimum}
Shiqi Shen, Yong Cheng, Zhongjun He, Wei He, Hua Wu, Maosong Sun, and Yang Liu.
  2016.
\newblock \href {https://doi.org/10.18653/v1/P16-1159} {Minimum risk training
  for neural machine translation}.
\newblock In \emph{Proceedings of the 54th Annual Meeting of the Association
  for Computational Linguistics (Volume 1: Long Papers)}, pages 1683--1692,
  Berlin, Germany. Association for Computational Linguistics.

\bibitem[{Sperber and Paulik(2020)}]{sperber-paulik-2020-speech}
Matthias Sperber and Matthias Paulik. 2020.
\newblock \href {https://doi.org/10.18653/v1/2020.acl-main.661} {Speech
  translation and the end-to-end promise: Taking stock of where we are}.
\newblock In \emph{Proceedings of the 58th Annual Meeting of the Association
  for Computational Linguistics}, pages 7409--7421, Online. Association for
  Computational Linguistics.

\bibitem[{Stoian et~al.(2020)Stoian, Bansal, and
  Goldwater}]{stoian2020analyzing}
Mihaela~C. Stoian, Sameer Bansal, and Sharon Goldwater. 2020.
\newblock Analyzing asr pretraining for low-resource speech-to-text
  translation.
\newblock In \emph{ICASSP 2020-2020 IEEE International Conference on Acoustics,
  Speech and Signal Processing (ICASSP)}, pages 7909--7913. IEEE.

\bibitem[{Tang et~al.(2022)Tang, Gong, Dong, Wang, Hsu, Gu, Baevski, Li,
  Mohamed, Auli, and Pino}]{tang-etal-2022-unified}
Yun Tang, Hongyu Gong, Ning Dong, Changhan Wang, Wei-Ning Hsu, Jiatao Gu,
  Alexei Baevski, Xian Li, Abdelrahman Mohamed, Michael Auli, and Juan Pino.
  2022.
\newblock \href {https://doi.org/10.18653/v1/2022.acl-long.105} {Unified
  speech-text pre-training for speech translation and recognition}.
\newblock In \emph{Proceedings of the 60th Annual Meeting of the Association
  for Computational Linguistics (Volume 1: Long Papers)}, pages 1488--1499,
  Dublin, Ireland. Association for Computational Linguistics.

\bibitem[{Tang et~al.(2021{\natexlab{a}})Tang, Pino, Li, Wang, and
  Genzel}]{tang-etal-2021-improving}
Yun Tang, Juan Pino, Xian Li, Changhan Wang, and Dmitriy Genzel.
  2021{\natexlab{a}}.
\newblock \href {https://doi.org/10.18653/v1/2021.acl-long.328} {Improving
  speech translation by understanding and learning from the auxiliary text
  translation task}.
\newblock In \emph{Proceedings of the 59th Annual Meeting of the Association
  for Computational Linguistics and the 11th International Joint Conference on
  Natural Language Processing (Volume 1: Long Papers)}, pages 4252--4261,
  Online. Association for Computational Linguistics.

\bibitem[{Tang et~al.(2021{\natexlab{b}})Tang, Pino, Wang, Ma, and
  Genzel}]{generalmtl}
Yun Tang, Juan Pino, Changhan Wang, Xutai Ma, and Dmitriy Genzel.
  2021{\natexlab{b}}.
\newblock \href {https://doi.org/10.1109/ICASSP39728.2021.9415058} {A general
  multi-task learning framework to leverage text data for speech to text
  tasks}.
\newblock In \emph{ICASSP 2021 - 2021 IEEE International Conference on
  Acoustics, Speech and Signal Processing (ICASSP)}, pages 6209--6213.

\bibitem[{Valk and Alum{\"{a}}e(2021)}]{voxlingua107}
J{\"{o}}rgen Valk and Tanel Alum{\"{a}}e. 2021.
\newblock \href {https://doi.org/10.1109/SLT48900.2021.9383459}
  {{VOXLINGUA107:} {A} dataset for spoken language recognition}.
\newblock In \emph{{IEEE} Spoken Language Technology Workshop, {SLT} 2021,
  Shenzhen, China, January 19-22, 2021}, pages 652--658. {IEEE}.

\bibitem[{Vaswani et~al.(2017)Vaswani, Shazeer, Parmar, Uszkoreit, Jones,
  Gomez, Kaiser, and Polosukhin}]{transformer}
Ashish Vaswani, Noam Shazeer, Niki Parmar, Jakob Uszkoreit, Llion Jones,
  Aidan~N Gomez, \L~ukasz Kaiser, and Illia Polosukhin. 2017.
\newblock \href
  {https://proceedings.neurips.cc/paper/2017/file/3f5ee243547dee91fbd053c1c4a845aa-Paper.pdf}
  {Attention is all you need}.
\newblock In \emph{Advances in Neural Information Processing Systems},
  volume~30. Curran Associates, Inc.

\bibitem[{Wang et~al.(2021{\natexlab{a}})Wang, Riviere, Lee, Wu, Talnikar,
  Haziza, Williamson, Pino, and Dupoux}]{wang-etal-2021-voxpopuli}
Changhan Wang, Morgane Riviere, Ann Lee, Anne Wu, Chaitanya Talnikar, Daniel
  Haziza, Mary Williamson, Juan Pino, and Emmanuel Dupoux. 2021{\natexlab{a}}.
\newblock \href {https://doi.org/10.18653/v1/2021.acl-long.80} {{V}ox{P}opuli:
  A large-scale multilingual speech corpus for representation learning,
  semi-supervised learning and interpretation}.
\newblock In \emph{Proceedings of the 59th Annual Meeting of the Association
  for Computational Linguistics and the 11th International Joint Conference on
  Natural Language Processing (Volume 1: Long Papers)}, pages 993--1003,
  Online. Association for Computational Linguistics.

\bibitem[{Wang et~al.(2021{\natexlab{b}})Wang, Wu, Pino, Baevski, Auli, and
  Conneau}]{DBLP:conf/interspeech/WangWPBAC21}
Changhan Wang, Anne Wu, Juan Pino, Alexei Baevski, Michael Auli, and Alexis
  Conneau. 2021{\natexlab{b}}.
\newblock \href {https://doi.org/10.21437/Interspeech.2021-1912} {Large-scale
  self- and semi-supervised learning for speech translation}.
\newblock In \emph{Interspeech 2021, 22nd Annual Conference of the
  International Speech Communication Association, Brno, Czechia, 30 August - 3
  September 2021}, pages 2242--2246. {ISCA}.

\bibitem[{Wang et~al.(2020{\natexlab{a}})Wang, Wu, and Pino}]{covost2}
Changhan Wang, Anne Wu, and Juan~Miguel Pino. 2020{\natexlab{a}}.
\newblock \href {http://arxiv.org/abs/2007.10310} {Covost 2: {A} massively
  multilingual speech-to-text translation corpus}.
\newblock \emph{CoRR}, abs/2007.10310.

\bibitem[{Wang and Sennrich(2020)}]{wang-sennrich-2020-exposure}
Chaojun Wang and Rico Sennrich. 2020.
\newblock \href {https://doi.org/10.18653/v1/2020.acl-main.326} {On exposure
  bias, hallucination and domain shift in neural machine translation}.
\newblock In \emph{Proceedings of the 58th Annual Meeting of the Association
  for Computational Linguistics}, pages 3544--3552, Online. Association for
  Computational Linguistics.

\bibitem[{Wang et~al.(2020{\natexlab{b}})Wang, Wu, Liu, Yang, and
  Zhou}]{wang2020bridging}
Chengyi Wang, Yu~Wu, Shujie Liu, Zhenglu Yang, and Ming Zhou.
  2020{\natexlab{b}}.
\newblock Bridging the gap between pre-training and fine-tuning for end-to-end
  speech translation.
\newblock In \emph{Proc. of AAAI}, volume~34, pages 9161--9168.

\bibitem[{Wang et~al.(2020{\natexlab{c}})Wang, Wu, Liu, Zhou, and
  Yang}]{wang-etal-2020-curriculum}
Chengyi Wang, Yu~Wu, Shujie Liu, Ming Zhou, and Zhenglu Yang.
  2020{\natexlab{c}}.
\newblock \href {https://doi.org/10.18653/v1/2020.acl-main.344} {Curriculum
  pre-training for end-to-end speech translation}.
\newblock In \emph{Proceedings of the 58th Annual Meeting of the Association
  for Computational Linguistics}, pages 3728--3738, Online. Association for
  Computational Linguistics.

\bibitem[{Weiss et~al.(2017)Weiss, Chorowski, Jaitly, Wu, and
  Chen}]{weiss2017sequence}
Ron~J. Weiss, Jan Chorowski, Navdeep Jaitly, Yonghui Wu, and Zhifeng Chen.
  2017.
\newblock Sequence-to-sequence models can directly translate foreign speech.
\newblock In \emph{{Proc. of INTERSPEECH}}, pages 2625--2629.

\bibitem[{Wu et~al.(2018)Wu, Tian, Qin, Lai, and Liu}]{wu-etal-2018-study}
Lijun Wu, Fei Tian, Tao Qin, Jianhuang Lai, and Tie-Yan Liu. 2018.
\newblock \href {https://doi.org/10.18653/v1/D18-1397} {A study of
  reinforcement learning for neural machine translation}.
\newblock In \emph{Proceedings of the 2018 Conference on Empirical Methods in
  Natural Language Processing}, pages 3612--3621, Brussels, Belgium.
  Association for Computational Linguistics.

\bibitem[{Xu et~al.(2021{\natexlab{a}})Xu, Hu, Li, Zhang, Huang, Ju, Xiao, and
  Zhu}]{xu-etal-2021-stacked}
Chen Xu, Bojie Hu, Yanyang Li, Yuhao Zhang, Shen Huang, Qi~Ju, Tong Xiao, and
  Jingbo Zhu. 2021{\natexlab{a}}.
\newblock \href {https://doi.org/10.18653/v1/2021.acl-long.204} {Stacked
  acoustic-and-textual encoding: Integrating the pre-trained models into speech
  translation encoders}.
\newblock In \emph{Proceedings of the 59th Annual Meeting of the Association
  for Computational Linguistics and the 11th International Joint Conference on
  Natural Language Processing (Volume 1: Long Papers)}, pages 2619--2630,
  Online. Association for Computational Linguistics.

\bibitem[{Xu et~al.(2021{\natexlab{b}})Xu, Liu, Meng, Zhang, Xu, and
  Zhou}]{xu-etal-2021-bilingual}
Yangyifan Xu, Yijin Liu, Fandong Meng, Jiajun Zhang, Jinan Xu, and Jie Zhou.
  2021{\natexlab{b}}.
\newblock \href {https://doi.org/10.18653/v1/2021.acl-short.65} {Bilingual
  mutual information based adaptive training for neural machine translation}.
\newblock In \emph{Proceedings of the 59th Annual Meeting of the Association
  for Computational Linguistics and the 11th International Joint Conference on
  Natural Language Processing (Volume 2: Short Papers)}, pages 511--516,
  Online. Association for Computational Linguistics.

\bibitem[{Ye et~al.(2021)Ye, Wang, and Li}]{ye2021end}
Rong Ye, Mingxuan Wang, and Lei Li. 2021.
\newblock \href
  {https://www.isca-speech.org/archive/pdfs/interspeech_2021/ye21_interspeech.pdf}
  {End-to-end speech translation via cross-modal progressive training}.
\newblock In \emph{Proc. of INTERSPEECH}.

\bibitem[{Ye et~al.(2022)Ye, Wang, and Li}]{ye-etal-2022-cross}
Rong Ye, Mingxuan Wang, and Lei Li. 2022.
\newblock \href {https://doi.org/10.18653/v1/2022.naacl-main.376} {Cross-modal
  contrastive learning for speech translation}.
\newblock In \emph{Proceedings of the 2022 Conference of the North American
  Chapter of the Association for Computational Linguistics: Human Language
  Technologies}, pages 5099--5113, Seattle, United States. Association for
  Computational Linguistics.

\bibitem[{Zhang et~al.(2020)Zhang, Williams, Titov, and
  Sennrich}]{zhang-etal-2020-improving}
Biao Zhang, Philip Williams, Ivan Titov, and Rico Sennrich. 2020.
\newblock \href {https://doi.org/10.18653/v1/2020.acl-main.148} {Improving
  massively multilingual neural machine translation and zero-shot translation}.
\newblock In \emph{Proceedings of the 58th Annual Meeting of the Association
  for Computational Linguistics}, pages 1628--1639, Online. Association for
  Computational Linguistics.

\bibitem[{Zhang et~al.(2022{\natexlab{a}})Zhang, Liu, Meng, Chen, Xu, Liu, and
  Zhou}]{zhang-etal-2022-conditional}
Songming Zhang, Yijin Liu, Fandong Meng, Yufeng Chen, Jinan Xu, Jian Liu, and
  Jie Zhou. 2022{\natexlab{a}}.
\newblock \href {https://doi.org/10.18653/v1/2022.acl-long.169} {Conditional
  bilingual mutual information based adaptive training for neural machine
  translation}.
\newblock In \emph{Proceedings of the 60th Annual Meeting of the Association
  for Computational Linguistics (Volume 1: Long Papers)}, pages 2377--2389,
  Dublin, Ireland. Association for Computational Linguistics.

\bibitem[{Zhang et~al.(2019)Zhang, Feng, Meng, You, and
  Liu}]{zhang-etal-2019-bridging}
Wen Zhang, Yang Feng, Fandong Meng, Di~You, and Qun Liu. 2019.
\newblock \href {https://doi.org/10.18653/v1/P19-1426} {Bridging the gap
  between training and inference for neural machine translation}.
\newblock In \emph{Proceedings of the 57th Annual Meeting of the Association
  for Computational Linguistics}, pages 4334--4343, Florence, Italy.
  Association for Computational Linguistics.

\bibitem[{Zhang et~al.(2022{\natexlab{b}})Zhang, Zhou, Ao, Liu, Dai, Li, and
  Wei}]{zhang2022speechut}
Ziqiang Zhang, Long Zhou, Junyi Ao, Shujie Liu, Lirong Dai, Jinyu Li, and Furu
  Wei. 2022{\natexlab{b}}.
\newblock Speechut: Bridging speech and text with hidden-unit for
  encoder-decoder based speech-text pre-training.
\newblock In \emph{Proceedings of the 2022 Conference on Empirical Methods in
  Natural Language Processing}, Online and Abu Dhabi.

\bibitem[{Zheng et~al.(2021)Zheng, Chen, Ma, and Huang}]{zheng2021fused}
Renjie Zheng, Junkun Chen, Mingbo Ma, and Liang Huang. 2021.
\newblock \href {http://arxiv.org/abs/2102.05766} {Fused acoustic and text
  encoding for multimodal bilingual pretraining and speech translation}.
\newblock In \emph{Proc. of ICML}.

\bibitem[{Zhou et~al.(2023)Zhou, Fang, and Feng}]{zhou-etal-2023-cmot}
Yan Zhou, Qingkai Fang, and Yang Feng. 2023.
\newblock {CMOT}: Cross-modal mixup via optimal transport for speech
  translation.
\newblock In \emph{Proceedings of the 61st Annual Meeting of the Association
  for Computational Linguistics}.

\end{thebibliography}
